\documentclass[11pt]{article}

\usepackage[final]{acl2026}
\usepackage{times}
\usepackage{latexsym}
\usepackage[T1]{fontenc}
\usepackage[utf8]{inputenc}
\usepackage{microtype}
\usepackage{inconsolata}
\usepackage{graphicx}
\usepackage{xspace}

\usepackage{booktabs}
\usepackage{booktabs}
\usepackage{multirow}
\usepackage{bbm}

\newcommand{\eg}{\hbox{\emph{e.g.},}\xspace}
\newcommand{\ie}{\hbox{\emph{i.e.},}\xspace}

\usepackage{pifont}
\usepackage[table]{xcolor}
\usepackage{xcolor}
\definecolor{my_green}{RGB}{40,154,121}
\definecolor{my_red}{RGB}{176,46,46}

\newcommand{\errormark}{\textcolor{my_red}{\ding{56}}}

\usepackage{amsmath}

\definecolor{bestcolor}{RGB}{242,167,167}
\definecolor{secondcolor}{RGB}{255,210,210}
\definecolor{thirdcolor}{RGB}{255,240,240}

\usepackage{algorithm}
\usepackage{algpseudocode}

\usepackage{tcolorbox}
\tcbuselibrary{breakable,skins,theorems}

\usepackage{enumitem}

\usepackage[utf8]{inputenc}
\usepackage[T1]{fontenc}
\usepackage{url}
\usepackage{
    amsmath
}
\usepackage{
    amssymb
}

\usepackage[dvipsnames]{
    xcolor
}
\usepackage{tcolorbox}
\tcbuselibrary{most}
\usepackage{listings}
\usepackage{caption}

\definecolor{chart}{HTML}{1f77b4}
\definecolor{arxiv}{HTML}{b31a1a}
\definecolor{MyPurple}{HTML}{8856A7}
\definecolor{MyOrange}{HTML}{D95F02}
\definecolor{MyGreen}{HTML}{66A61E}

\newtcolorbox{custombox}[1][]{ colback=chart!5!white,
colframe=chart,
floatplacement=floating,
title=\centering #1
}

\newtcolorbox{format_error}[1][]{ enhanced,
colframe=MyPurple,
colback=MyPurple!10!white,
sharp corners,
boxsep=0pt,
left=6pt,
right=6pt,
top=6pt,
bottom=6pt,
boxrule=0pt,
leftrule=4pt,
#1
}

\newtcolorbox{redundancy_error}[1][]{ enhanced,
colframe=MyOrange,
colback=MyOrange!10!white,
sharp corners,
boxsep=0pt,
left=6pt,
right=6pt,
top=6pt,
bottom=6pt,
boxrule=0pt,
leftrule=4pt,
#1 }

\definecolor{KeyStepGreen}{HTML}{1B6B4E}

\newtcolorbox{key_step_analysis}[1][]{%
enhanced,
colframe=KeyStepGreen!80!black,
colback=KeyStepGreen!7!white,
sharp corners,
boxsep=0pt,
left=6pt,
right=6pt,
top=6pt,
bottom=6pt,
boxrule=0pt,
leftrule=4pt,
title={#1},
fonttitle=\bfseries\color{KeyStepGreen!80!black},
}

\newtcolorbox{evaluation_format_error}[1][]{%
enhanced,
colframe=BrickRed!80!black,
colback=BrickRed!10!white,
sharp corners,
boxsep=0pt,
left=6pt,
right=6pt,
top=6pt,
bottom=6pt,
boxrule=0pt,
leftrule=4pt,
title={#1},
fonttitle=\bfseries\color{BrickRed!80!black},
before skip=10pt plus 3pt,
after skip=10pt plus 3pt,
#1
}

\usepackage{fontawesome5}

\newcommand{\homepage}{\raisebox{-1.5pt}{\includegraphics[height=1.2em]{
  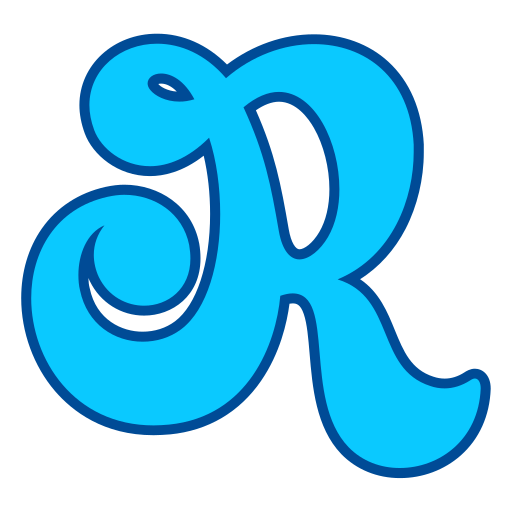
}}}
\newcommand{\github}{\raisebox{-1.5pt}{\includegraphics[height=1em]{
  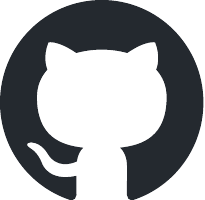
}}}
\newcommand{\huggingface}{%
\raisebox{-1.5pt}{\includegraphics[height=1em]{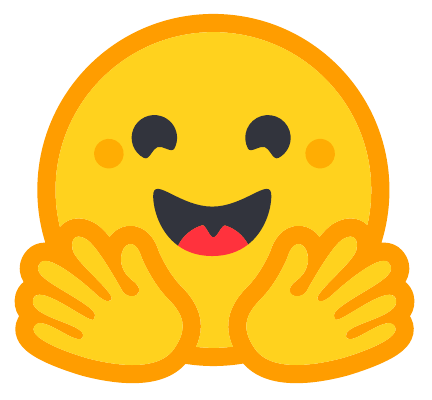}}%
}

\title{RoadMapper: A Multi-Agent System for \\
Roadmap Generation of Solving Complex Research Problems}

\author{\textbf{Jiacheng Liu}\textsuperscript{\rm 1} \quad \textbf{Zichen Tang}\textsuperscript{\rm 1}
\quad \textbf{Zhongjun Yang}\textsuperscript{\rm 1} \quad \textbf{Xinyi Hu}\textsuperscript{\rm 1}
\quad \textbf{Xueyuan Lin}\textsuperscript{\rm 2,3,4}\\
\textbf{Linwei Jia}\textsuperscript{\rm 1} \quad \textbf{Ruofei Bai}\textsuperscript{\rm 1}
\quad \textbf{Rongjin Li}\textsuperscript{\rm 1} \quad \textbf{Shiyao Peng}\textsuperscript{\rm 1}
\quad \textbf{Haocheng Gao}\textsuperscript{\rm 1} \quad \textbf{Haihong E}\textsuperscript{\rm 1}\thanks{Corresponding author.}\\
\textsuperscript{\rm 1}Beijing University of Posts and Telecommunications\\
\textsuperscript{\rm 2}The Hong Kong University of Science and Technology (Guangzhou)\\
\textsuperscript{\rm 3}IDEA Research \quad \textsuperscript{\rm 4}Hithink RoyalFlush
Information Network Co., Ltd.\\
\homepage~~\href{https://bupt-reasoning-lab.github.io/RoadMapper}{\texttt{bupt-reasoning-lab.github.io/RoadMapper}}\\
\github~~\href{https://github.com/BUPT-Reasoning-Lab/RoadMapper}{\texttt{BUPT-Reasoning-Lab/RoadMapper}}
\quad \huggingface~~\href{https://huggingface.co/datasets/BUPT-Reasoning-Lab/RoadMapper}{\texttt{BUPT-Reasoning-Lab/RoadMapper}}
}

\begin{document}
  \maketitle

  \begin{abstract}
    People commonly leverage structured content to accelerate knowledge acquisition
    and research problem solving. Among these, roadmaps guide researchers through
    hierarchical subtasks to solve complex research problems step by step.
    Despite progress in structured content generation, the \textbf{roadmap
    generation task} has remained unexplored. To bridge this gap, we introduce
    \textbf{RoadMap}, a novel benchmark designed to evaluate the ability of
    large language models (LLMs) to construct high-quality roadmaps for solving
    complex research problems. Based on this, we identify three limitations of
    LLMs: (1) \textit{lack of professional knowledge}, (2) \textit{unreasonable
    task decomposition}, and (3) \textit{disordered logical relationships}. To address
    these challenges, we propose \textbf{RoadMapper}, an LLM-based multi-agent
    system that decomposes the research roadmap generation task into three key
    stages (\ie{ initial generation, knowledge augmentation, and iterative ``critique-revise-evaluate''}).
    Extensive experiments demonstrate that RoadMapper can improve LLMs' ability for
    roadmap generation, while enhancing average performance by more than \textbf{8\%}
    and \textbf{saving 84\% of the time} required by human experts, highlighting
    its effectiveness and application potential.
\end{abstract}
  \section{Introduction}
\begin{figure}[t!]
    \centering
    \includegraphics[width=\columnwidth]{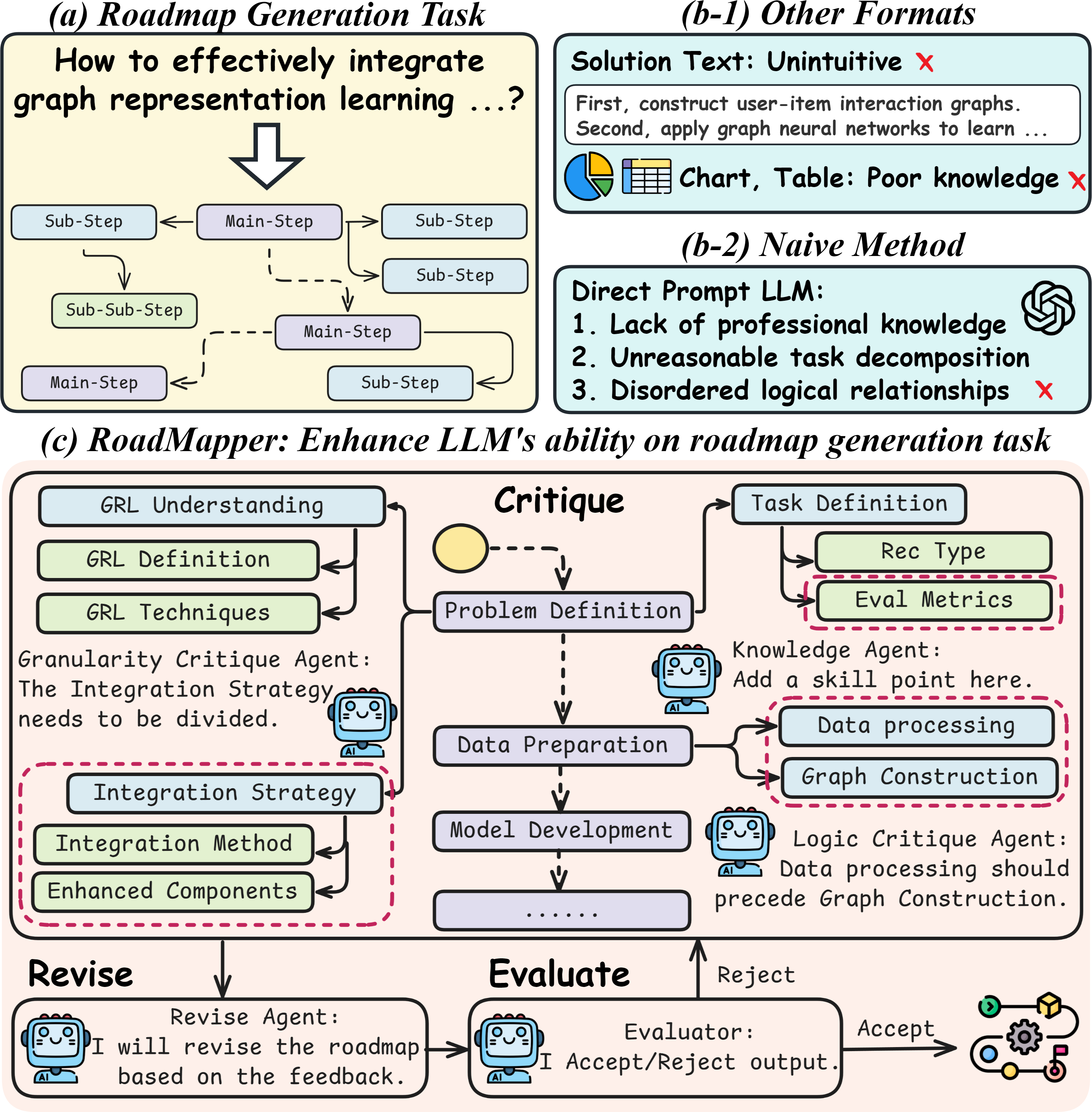}
    \caption{We propose the \textit{roadmap generation task}. Prior answer
    formats and methods face multiple challenges when solving complex research
    problems, but \textbf{RoadMapper} effectively addresses these challenges
    through an iterative ``critique-revise-evaluate'' process.}
    \label{fig:01_ourwork}
\end{figure}
\begin{table*}
    [t!]
    \centering
    \resizebox{\linewidth}{!}{
    \begin{tabular}{lccccccccc}
        \toprule \multirow{2.5}{*}{\textbf{Research}}                               & \multicolumn{2}{c}{\textbf{Basic Statistics}} & \multicolumn{3}{c}{\textbf{Field and Knowledge}} & \textbf{Professional Depth} & \multicolumn{2}{c}{\textbf{Guidance}} \\
        \cmidrule(lr){2-3} \cmidrule(lr){4-6} \cmidrule(lr){7-7} \cmidrule(lr){8-9} & \textbf{\# Samples}                           & \textbf{Method}                                  & \textbf{\# Fields}              & \textbf{\# Types}                        & \textbf{Language} & \textbf{Material}      & \textbf{Output Format}  & \textbf{Application} \\
        \midrule Seq2Seqset~~\citep{li-etal-2023-sequence-sequence}                 & 4,855                                         & Seq2Seq                                          & 1                           & 1                                    & EN                & News Reports            & Table                   & I                    \\
        Text-Tuple-Table~~\citep{deng-etal-2024-text}                               & 3,771                                         & Prompting                                        & 1                           & 1                                    & EN                & Live Commentaries        & Table                   & I                    \\
        End-to-End Parsing~~\citep{bhatt-etal-2024-end}                             & 10,300                                        & End2End                                          & 1                           & 1                                    & EN                & Cooking Recipes        & Flow Graph              & G                    \\
        StructSum~~\citep{jain-etal-2024-structsum}                                 & 200                                           & Prompting                                        & N/A                  & 1                                    & EN                & Wiki Pages             & Table+Mindmap           & I                    \\
        WHPG~~\citep{ren-etal-2023-constructing}                                    & 283                                           & End2End                                          & N/A                  & 1                                    & EN                & Wiki Pages             & Procedural Graph        & G                    \\
        EMGN~~\citep{hu-etal-2021-efficient}                                        & 44,585                                        & End2End+RL                                       & N/A                  & 1                                    & EN                & News Articles          & Mindmap                 & I                    \\
        COMET~~\citep{bosselut-etal-2019-comet}                                     & 977,000                                       & End2End                                          & N/A                  & N/A                           & EN                & N/A             & Knowledge Graph         & G                    \\
        DeepSolution~~\citep{li-etal-2025-deepsolution}                             & 3,024                                         & RAG                                              & 8                           & 1                                    & EN                & Technical Reports      & Solution Text           & G                    \\
        METAL~~\citep{li-etal-2025-metal}                                           & 1,000                                         & MAS                                              & 2                           & 2                                    & EN                & Charts+Instructions    & Chart Code              & I                    \\
        \midrule \textbf{RoadMapper (ours)}                                                & \textbf{1,705}                                & \textbf{MAS}                                     & \textbf{10}                 & \textbf{5}                           & \textbf{EN+CN}    & \textbf{Dissertations} & \textbf{Roadmap (ours)} & \textbf{G}           \\
        \bottomrule
    \end{tabular}
    }
    \caption{Comparison between \textbf{RoadMapper} and related research on structured
    content generation. \textbf{I}:
    Information Display, \textbf{G}: Guidance. RoadMapper offers three key
    advantages: (1) \textit{Field and knowledge coverage} are ensured by
    multiple fields, diverse types, and bilingual support; (2) \textit{Professional
    depth} is ensured by dissertations written by Master's and Ph.D. students; (3) \textit{Step-by-step
    guidance} is provided through a structured roadmap format.}
    \label{tab:comparison}
\end{table*}
Designing a structured roadmap for complex research problems is an important
task in scientific research and education~\citep{burian2010research}. Solving
these research problems often faces challenges such as a wide range of knowledge
and rapid technological iteration. A meticulously designed roadmap can break down
these research problems into multiple subtasks, thus improve the efficiency and
quality of solutions~\citep{sorensen2024position}.

Currently, designing research roadmaps heavily relies on human experts who
create them by consulting professional knowledge, meticulously designing, and
iteratively conducting reviews. However, this manual process is both time-consuming
and resource-demanding. Fortunately, recent advances in LLMs~\citep{park-kim-2025-understanding}
present an opportunity to develop an automatic system to generate high-quality roadmaps.

However, as shown in Table~\ref{tab:comparison}, current research has not sufficiently
explored the roadmap generation task, usually suffering from three limitations:

\begin{itemize}[leftmargin=*]
    \item \textbf{Limited Field and Knowledge Coverage.} Complex research
        problems typically span multiple professional fields, and their solutions
        similarly demand the integration of expertise from diverse disciplines.
        However, existing research focuses primarily on a few specific fields~\citep{li-etal-2023-sequence-sequence, deng-etal-2024-text}.

    \item \textbf{Insufficient Professional Depth.} Solving complex research
        problems requires integrating professional knowledge for deep analysis. However,
        existing research usually focuses on shallow tasks, such as extracting cooking
        flow graphs from recipes~\citep{bhatt-etal-2024-end} or converting text
        to mindmaps~\citep{hu-etal-2021-efficient}.

    \item \textbf{Lack of Step-by-Step Guidance.} Logically coherent roadmaps
        help guide people to solve complex research problems step by step. However,
        existing research in formats such as tables and knowledge graphs is
        mainly designed for information display rather than offering guidance~\citep{li-etal-2023-sequence-sequence, bosselut-etal-2019-comet}.
\end{itemize}

To bridge this gap, we construct \textbf{RoadMap}, a novel benchmark focusing on
evaluating LLMs' capabilities for generating high-quality roadmaps of solving research
problems, \textbf{covering 10 research fields, 5 research types, and 2 languages}
(\ie{ English and Chinese}). RoadMap comprises two sub-components: (1) \textbf{\textit{Skill-Repo}}:
This component consists of 8,436 professional skill points, each of which contains
a name, a detailed description, and a problem it solves, serving to provide LLMs
with extensive professional knowledge; (2) \textbf{\textit{Golden-Roadmap}}: This
component consists of 1,705 complex research problems, with each accompanied by
a golden roadmap that was meticulously designed by experts and serves as a
reference for evaluating roadmaps from automatic systems.

We conduct extensive experiments on RoadMap and find that direct or repeated
prompting of LLMs faces three main challenges: \textit{\textbf{Q1}: lack of
professional knowledge, \textbf{Q2}: unreasonable task decomposition, and
\textbf{Q3}: disordered logical relationships}, exceeding the capabilities of a single
model.

To address these challenges, we propose our \textbf{RoadMapper}, a multi-agent
system that emulates the manual process of human experts and decomposes the
roadmap generation task into three stages: initial generation, knowledge
augmentation, and iterative ``critique-revise-evaluate''. Specifically, \textbf{(1)}
the \textit{Init agent} generates an initial roadmap based on the research problem;
\textbf{(2)} the \textit{Knowledge agent} augments the initial roadmap by
knowledge from Skill-Repo, addressing Q1; \textbf{(3)} the \textit{Granularity
Critique agent} analyzes the decomposition granularity of sub-task nodes and
outputs revision suggestions, addressing Q2; \textbf{(4)} the \textit{Logic
Critique agent} analyzes the logical relationships between sub-task nodes and
outputs revision suggestions, addressing Q3; \textbf{(5)} the \textit{Revise
agent} implements revisions to the roadmap based on suggestions from the
critique agents; \textbf{(6)} the \textit{Evaluate agent} evaluates the roadmap quality
to determine whether to output as final result. To keep the evaluation aligned
with human expert, we train the Evaluate agent using \textbf{Direct Preference
Optimization (DPO) algorithm}, leveraging data from RoadMap and its construction
process. These LLM-driven agents will collaborate iteratively until the roadmap
reaches expected quality or the maximum number of iterations.

Finally, we conduct extensive experiments on RoadMap and evaluate using \textbf{four
novel metrics} from both structure (\ie{ \textbf{DegreeScore}, \textbf{DepthScore}})
and content (\ie{ \textbf{StepScore}, \textbf{LogicScore}}). The results demonstrate
that RoadMapper improves LLMs' performance on roadmap generation tasks: \textbf{addressing
Q1 by improving StepScore by 8.24, Q2 by improving structure metrics by 7.04,
and Q3 by improving LogicScore by 7.79}. Meanwhile, RoadMapper can reduce
designing time by more than \textbf{84\%} compared to human experts.

\begin{figure*}[t!]
    \centering
    \includegraphics[width=\textwidth]{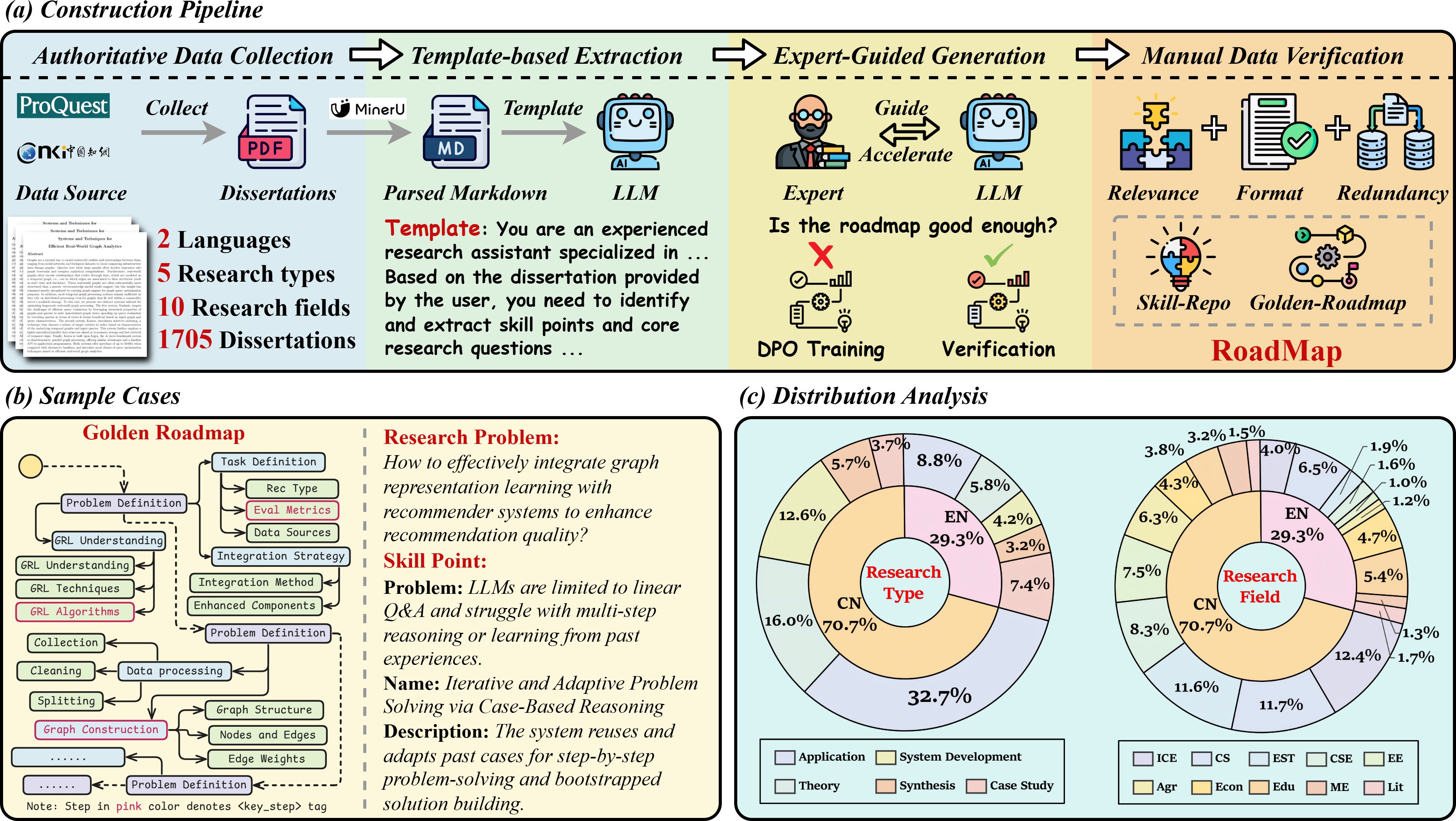}
    \caption{ Overview of \textbf{RoadMap}. \textbf{ICE}: Information and Communication
    Engineering, \textbf{CS}: Computer Science, \textbf{EST}: Electronic Science
    and Technology, \textbf{CSE}: Control Science and Engineering, \textbf{EE}: Electrical
    Engineering, \textbf{Agr}: Agronomy, \textbf{Econ}: Economics, \textbf{Edu}:
    Education, \textbf{ME}: Mechanical Engineering, \textbf{Lit}: Literature.}
    \label{fig:roadmap_bench}
\end{figure*}
Our main contributions are as follows:

\begin{itemize}[leftmargin=*]
    \item \textbf{We define the roadmap generation task and construct RoadMap}, providing
        extensive professional knowledge for roadmap generation systems and effectively
        evaluating the performance of systems on this task.

    \item \textbf{We propose RoadMapper}, a multi-agent system that emulates human
        experts to decompose the generation task into three iterative stages which
        are handled by six LLM-driven agents aligned with experts via DPO training.

    \item \textbf{We introduce four novel evaluation metrics and conduct
        extensive experiments}, with results demonstrating that RoadMapper can significantly
        address challenges faced by LLMs and generate well-designed roadmaps with
        strong structure and content efficiently.
\end{itemize}
  \section{Task Definition}
\label{sec:task_definition}

Mapping $\mathcal{F}$ describes the \textbf{roadmap generation task}:
\begin{equation}
    \mathcal{F}:\ x_{\text{problem}}\rightarrow y_{\text{roadmap}}.
\end{equation}

That is, for a given research problem $x_{\text{problem}}$, the target is to
output the corresponding $y_{\text{roadmap}}$, which is defined as a logically tree-like
roadmap guiding people to solve $x_{\text{problem}}$ step by step.

We represent roadmaps using Markdown documents that adhere to the following structural
rules: (1) Each line represents a task node, including \texttt{level}, \texttt{index},
and \texttt{title}; (2) The \texttt{level} is consistent with Markdown's heading
syntax, composed of several \# symbols; (3) The \texttt{index} satisfies the regular
expression \texttt{(\textbackslash d+(?:\textbackslash.\textbackslash d+)*)}, such
as \texttt{1.3.2}; (4) The \texttt{title} is enclosed in square brackets.
  \section{RoadMap Benchmark}
\label{sec:benchmark}

We introduce the construction of RoadMap in this section, with its overview in
Figure~\ref{fig:roadmap_bench}, key statistics in Table~\ref{tab:statistics_of_benchmark},
and additional details in Appendix~\ref{app:details_roadmap}.

\subsection{Authoritative Data Collection}
\label{data_collection}

We collect dissertations from ProQuest and CNKI, adopting the following
strategies: (1) published since 2018, (2) authored by graduate students, and (3)
from universities with strong academic programs. These dissertations offer three
key advantages: \textbf{\textit{Strong Relevance}}. Dissertations address cutting-edge
and complex research problems that align with our focus; \textbf{\textit{Extensive
Professional Knowledge}}. Dissertations contain complete solutions, detailed processes
and comprehensive results suitable for knowledge extraction; \textbf{\textit{Authoritative
Content}}. Expert scrutiny ensures high-quality data with consistently rigorous
logic.

\subsection{Template-Based Extraction}
\label{sec:template_based_extraction} We first parse the PDF files into Markdown~\citep{wang2024mineruopensourcesolutionprecise}
and filter out irrelevant information. Then, we manually format a template for information
extraction via \texttt{Gemini 2.5 Flash}, based on generative information extraction~\citep{lu-etal-2022-unified,10.1007/s11704-024-40555-y,zhang-etal-2025-survey}.
The template requires the following information from each dissertation: (1)
\textbf{\textit{Core research problem}}, which is the main scientific or
technical problem the dissertation aims to solve; (2) \textbf{\textit{Skill
Points}}, which are the critical technologies used in the dissertation to address
the research problem, each including its name, a detailed description, and the
problem it solves. Finally, we save the results in JSON format.

\subsection{Expert-Guided Generation}
\label{sec:expert_guided_generation}

\begin{table}[t!]
    \centering
    \resizebox{\columnwidth}{!}{
    \begin{tabular}{lr}
        \toprule \textbf{Property}              & \textbf{Value}      \\
        \midrule \# Golden Roadmaps (EN/CN)     & 1,705 (500/1,205)   \\
        \# Skill Points (EN/CN)                 & 8,436 (2,493/5,943) \\
        \# Research Fields                      & 10                  \\
        \# Research Types                       & 5                   \\
        \midrule \# Avg. Key Steps              & 21.60               \\
        \# Avg. Depth (Knowledge Depth)         & 3.22                \\
        \# Avg. Out-Degree (Knowledge Breadth)  & 3.16                \\
        \# Avg. Leaf Nodes (Knowledge Richness) & 86.46               \\
        \bottomrule
    \end{tabular}
    }
    \caption{Key statistics of RoadMap.}
    \label{tab:statistics_of_benchmark}
\end{table}

We assemble experts from diverse research fields to develop a golden roadmap for
each research problem. These experts are allowed to leverage \texttt{Gemini 2.5
Flash} to accelerate development, while remaining primarily guided by their own experience.
Specifically, experts first synthesize the main framework of roadmaps based on their
domain knowledge and relevant dissertations. Subsequently, this framework is used
to prompt LLMs for generating initial drafts. Experts then iteratively guide the
model to refine problematic components until a high-quality roadmap is achieved.
Finally, experts annotate critical steps in the roadmap using \texttt{<key\_step>},
which are essential for experimental evaluation. Notably, problematic roadmaps
will not be discarded as they are valuable for DPO training.

\subsection{Manual Data Verification}
\label{sec:quality_assurance}

We implement rigorous verification processes to prevent errors from LLMs (\eg{ hallucinations})
and ensure benchmark quality~\citep{10.1145/3703155}: (1) \textbf{\textit{Relevance
Verification}}. Two experts cross-validate the model-extracted results to
determine if they faithfully match the corresponding dissertations. Any mismatched
content will be removed. In cases of disagreement, one additional expert will be
introduced for arbitration; (2) \textbf{\textit{Format Verification}}. We
develop an automated Python script to validate the formatting of roadmaps and
skill points against predefined requirements. Any incorrect content (about 19.3\%)
will be manually revised by experts; (3) \textbf{\textit{Redundancy Removal}}. We
use \texttt{Qwen3 Embedding 8B} to help identify skill points with high
similarity. Any duplicate contents (about 5.3\%) will be discarded or manually
merged by experts depending on the type of redundancy.
  \section{RoadMapper Methodology}
\label{sec:methodology}

\begin{figure*}[t!]
    \centering
    \includegraphics[width=\textwidth]{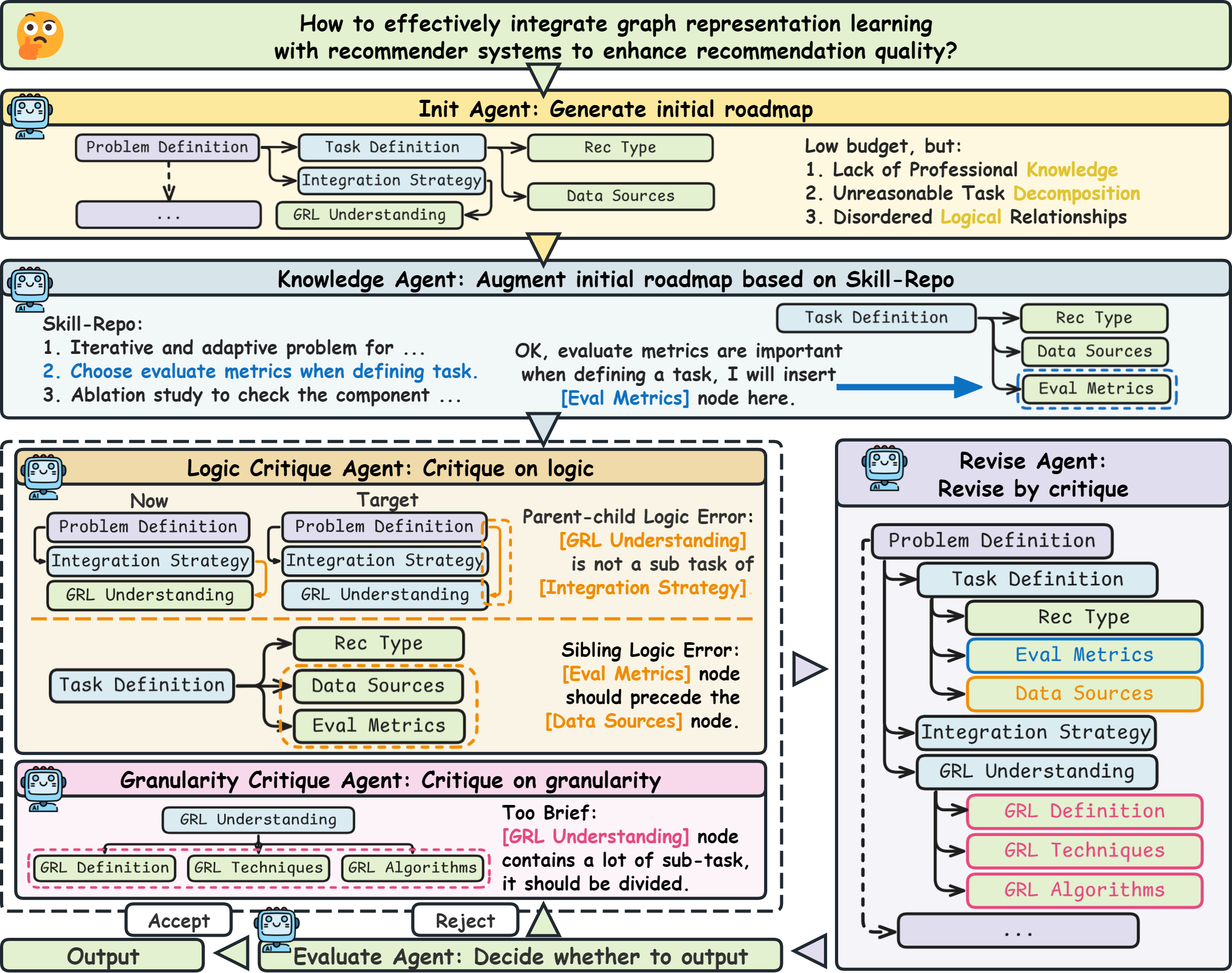}
    \caption{ Overview of \textbf{RoadMapper}, a multi-agent system consisting of
    six agents: Init Agent ($\mathcal{I}$), Knowledge Agent ($\mathcal{K}$),
    Logic Critique Agent ($\mathcal{L}$), Granularity Critique Agent ($\mathcal{G}$),
    Revise Agent ($\mathcal{R}$), and Evaluate Agent ($\mathcal{E}$).}
    \label{fig:RoadMapper}
\end{figure*}

We introduce the agents design of RoadMapper and the DPO training of the
Evaluate agent in this section, with additional details shown in Appendix~\ref{app:details_roadmapper}.

\subsection{Agents Design}
\label{sec:agents_design}

RoadMapper comprises six LLM agents prompted by carefully designed prompts, shown
in Figure~\ref{fig:RoadMapper}.

\subsubsection{Init Agent (\texorpdfstring{$\mathcal{I}$}{I})}
This agent is responsible for receiving research problems and generating an initial
roadmap draft:
\begin{equation}
    \mathcal{I}: y_{\text{initial}}= \mathcal{I}(x_{\text{problem}}).
\end{equation}

Here, $x_{\text{problem}}$ denotes the complex research problem and
$y_{\text{initial}}$ denotes the initial roadmap.

\subsubsection{Knowledge Agent (\texorpdfstring{$\mathcal{K}$}{K})}
This agent is responsible for augmenting the initial roadmap based on skill
points from Skill-Repo:
\begin{equation}
    \mathcal{K}: y_{0}= \mathcal{K}(knowledge, y_{initial}).
\end{equation}

Here, $knowledge$ denotes knowledge composed of the top-K skill points retrieved
from Skill-Repo via vector similarity, $y_{initial}$ denotes the initial roadmap,
and $y_{0}$ denotes the augmented roadmap.

\subsubsection{Logic Critique Agent (\texorpdfstring{$\mathcal{L}$}{L})}
This agent is responsible for critiquing the logic between nodes and outputting revision
suggestions:
\begin{equation}
    \mathcal{L}: LC_{t}= \mathcal{L}(y_{t}).
\end{equation}

Here, $y_{t}$ denotes the roadmap to be critiqued and $LC_{t}$ denotes the
logical revision suggestions.

We define two types of logical relationships: (1) \textit{\textbf{Parent-child
Logic}}: A child node must represent a direct refinement or an execution step of
its parent node's task; (2) \textit{\textbf{Sibling Logic}}: Sibling nodes of
the same parent must exhibit a parallel-progressive relationship among their
respective tasks.

\subsubsection{Granularity Critique Agent (\texorpdfstring{$\mathcal{G}$}{G})}
This agent is responsible for critiquing the granularity of nodes and outputting
revision suggestions:
\begin{equation}
    \mathcal{G}: GC_{t}= \mathcal{G}(y_{t}).
\end{equation}

Here, $y_{t}$ denotes the roadmap to be critiqued and $GC_{t}$ denotes the granularity
revision suggestions.

We define two types of inappropriate granularity: (1) \textit{\textbf{Too
Detailed}}: A node is split into an excessive number of trivial subtasks, introducing
unnecessary complexity and inefficiency; (2) \textit{\textbf{Too Brief}}: A node
remains information-dense and requires further decomposition, causing difficulty
in comprehension.

\subsubsection{Revise Agent (\texorpdfstring{$\mathcal{R}$}{R})}
This agent is responsible for revising the roadmap based on revision suggestions
from $\mathcal{L}$ and $\mathcal{G}$ agents:
\begin{equation}
    \mathcal{R}: y_{t+1}= \mathcal{R}(y_{t}, LC_{t}, GC_{t}).
\end{equation}

Here, $y_{t}$ denotes the roadmap to be revised, $LC_{t}$ denotes the logical revision
suggestions, $GC_{t}$ denotes the granularity revision suggestions, and $y_{t+1}$
denotes the revised roadmap.

\subsubsection{Evaluate Agent (\texorpdfstring{$\mathcal{E}$}{E})}
This agent is responsible for evaluating roadmaps and outputting objective evaluation
outcomes:
\begin{equation}
    \mathcal{E}: E_{t}= \mathcal{E}(y_{t+1}).
\end{equation}

Here, $y_{t+1}$ denotes the roadmap to be evaluated and $E_{t}$ denotes the
evaluation outcome.

$E_{t}$ comprises a score and a corresponding reason. The ``critique-revise-evaluate''
iteration will end if the score achieves a predefined \emph{passing score}.

\begin{table}[t]
    \centering
    \resizebox{\columnwidth}{!}{
    \begin{tabular}{rl}
        \toprule \multicolumn{2}{c}{\textbf{Algorithm 1 } Inference Procedure of RoadMapper} \\
        \midrule \textbf{Input:}                                                            & \begin{tabular}{@{}l@{}}$x_{\text{problem}}$: Research problem\end{tabular}        \\
        \addlinespace \textbf{Output:}                                                      & \begin{tabular}{@{}l@{}}$y^{*}$: Final refined output\end{tabular}                \\
        \midrule \textbf{1:}                                                                & $y_{\text{initial}}\leftarrow \mathcal{I}(x_{\text{problem}})$                     \\
        \textbf{2:}                                                                         & $y_{\text{knowledge}}\leftarrow \mathcal{K}(y_{\text{initial}}, \text{knowledge})$ \\
        \textbf{3:}                                                                         & $y_{0}\leftarrow y_{\text{knowledge}}, \quad t \leftarrow 0$                            \\
        \textbf{4:}                                                                         & \textbf{while} $t < T_{\max}$ \textbf{do}                                          \\
        \textbf{5:}                                                                         & \quad $LC_{t}\leftarrow \mathcal{L}(y_{t}), \quad GC_{t}\leftarrow \mathcal{G}(y_{t})$  \\
        \textbf{6:}                                                                         & \quad $y_{t+1}\leftarrow \mathcal{R}(y_{t}, LC_{t}, GC_{t})$                       \\
        \textbf{7:}                                                                         & \quad $E_{t}\leftarrow \mathcal{E}(y_{t+1})$                                       \\
        \textbf{8:}                                                                         & \quad \textbf{if} $E_{t}= \texttt{ACCEPT}$ \textbf{then}                             \\
        \textbf{9:}                                                                         & \quad\quad \textbf{return} $y_{t+1}$                                               \\
        \textbf{10:}                                                                        & \quad \textbf{else}                                                                \\
        \textbf{11:}                                                                        & \quad\quad $t \leftarrow t + 1$                                                    \\
        \textbf{12:}                                                                        & \quad \textbf{end if}                                                              \\
        \textbf{13:}                                                                        & \textbf{end while}                                                                 \\
        \textbf{14:}                                                                        & \textbf{return} $y_{t}$ as $y^{*}$                                                 \\
        \bottomrule
    \end{tabular}}
    \caption{Inference procedure of RoadMapper.}
    \label{tab:inference_procedure}
\end{table}
\subsection{DPO Training of Evaluate Agent}
\label{sec:dpo_training}

The reliability of agent $\mathcal{E}$ is critical to RoadMapper's efficiency and
roadmap quality, yet it remains susceptible to degradation from model biases. To
mitigate this issue, we employ DPO training to align $\mathcal{E}$ agent with
domain experts. Specifically, we use Qwen3 8B/14B/32B as backbone models due to
their demonstrated exceptional capabilities.

\paragraph{Preference Dataset Construction.}
We first collect roadmaps from RoadMap and its construction. For each roadmap
$x$, we employ \texttt{Qwen3-32B} to generate 10 evaluation candidates, using the
same prompt with $\mathcal{E}$ agent. Then, seven experts will vote to select samples:
(1) The candidate with \textbf{highest} number of votes is selected as the positive
sample $y_{w}$; (2) Crucially, we select the candidate receiving the \textbf{second-highest}
number of votes as the negative sample $y_{l}$ (rather than the lowest-ranked one)
because the second-best evaluations often highly resemble the optimal ones yet
exhibit subtle flaws, which enables the model to discern fine-grained expert preferences.
We also introduce format preference pairs to improve the formatting accuracy of
$\mathcal{E}$ agent. Finally, our preference dataset $\mathcal{D}$ contains 818
roadmaps with corresponding preference pairs.

\paragraph{Optimization Objective.}
We fine-tune the $\mathcal{E}$ agent, parameterized as $\pi_{\theta}$, to maximize
the relative log-likelihood of the preferred evaluation $y_{w}$ over the dispreferred
$y_{l}$, constrained by the reference model $\pi_{\text{ref}}$. The objective is
formulated as:
\begin{equation}
    \resizebox{0.89\linewidth}{!}{$\mathcal{L}_{\text{DPO}}= - \mathbb{E}_{\mathcal{D}}
    \left[ \log \sigma \left( \beta \log \frac{\pi_{\theta}(y_{w}| x) / \pi_{\text{ref}}(y_{w}|
    x)}{\pi_{\theta}(y_{l}| x) / \pi_{\text{ref}}(y_{l}| x)}\right) \right]$.}
\end{equation}

Here, $\mathcal{D}$ denotes the preference dataset, $\sigma$ denotes the logistic
sigmoid function, and $\beta$ is a hyperparameter regulating the deviation from
the reference policy. This optimization effectively aligns the agent's internal scoring
standards with expert judgment, ensuring the reliability and convergence of the
``critique-revise-evaluate'' loop.

\begin{table*}
        [!t]
        \centering
        \resizebox{\textwidth}{!}{
        \begin{tabular}{l|c|ccccc|ccccc|c}
                \toprule \multirow{2.5}{*}{\textbf{Base Model}}           & \multirow{2.5}{*}{\textbf{Method}} & \multicolumn{5}{c|}{\textbf{English}}    & \multicolumn{5}{c|}{\textbf{Chinese}} & \textbf{Overall }                      \\
                \cmidrule(lr){3-7}\cmidrule(lr){8-12}\cmidrule(lr){13-13} &                                    & \textbf{SS}                              & \textbf{LS}                           & \textbf{DegS}                         & \textbf{DepS}                         & \textbf{Avg.}                           & \textbf{SS}                             & \textbf{LS}                           & \textbf{DegS}                         & \textbf{DepS}                         & \textbf{Avg.}                         & \textbf{Avg.}                           \\
                \midrule

\multirow{3}{*}{Llama 3.1 8B}                   & DP                                 & 42.58                                    & 54.29                                 & 64.11                                 & 83.96                                 & 61.24                                   & 37.63                                   & 48.81                                 & 70.23                                 & 83.37                                 & 60.01                                 & 60.62                                   \\
                                                                          & BN                                 & \underline{44.24}                        & \underline{54.88}                     & \underline{64.23}                     & \underline{84.17}                     & \underline{61.88}                       & \underline{39.75}                       & \underline{48.97}                     & \underline{70.94}                     & \underline{83.66}                     & \underline{60.83}                     & \underline{61.36}                       \\
                                                                          & Ours                               & \textbf{49.98}                           & \textbf{62.23}                        & \textbf{71.95}                        & \textbf{86.57}                        & \textbf{67.68}                          & \textbf{48.51}                          & \textbf{58.79}                        & \textbf{76.36}                        & \textbf{88.15}                        & \textbf{67.95}                        & \textbf{67.82}                          \\
                \midrule

\multirow{3}{*}{Llama 4 Maverick}               & DP                                 & \underline{46.78}                        & \underline{56.91}                     & 61.92                                 & 86.97                                 & \underline{63.15}                       & 46.14                                   & 55.61                                 & 64.37                                 & 88.36                                 & 63.62                                 & 63.38                                   \\
                                                                          & BN                                 & 46.63                                    & 56.47                                 & \underline{61.95}                     & \underline{87.30}                     & 63.09                                   & \underline{46.29}                       & \underline{56.39}                     & \underline{64.44}                     & \underline{88.59}                     & \underline{63.93}                     & \underline{63.51}                       \\
                                                                          & Ours                               & \textbf{54.33}                           & \textbf{65.34}                        & \textbf{69.81}                        & \cellcolor{thirdcolor}\textbf{92.25}  & \textbf{70.43}                          & \textbf{53.11}                          & \textbf{64.35}                        & \textbf{74.32}                        & \cellcolor{thirdcolor}\textbf{93.30}  & \textbf{71.27}                        & \textbf{70.85}                          \\
                \midrule

\multirow{3}{*}{Qwen3-14B}                      & DP                                 & 54.76                                    & 62.64                                 & 73.56                                 & 86.04                                 & 69.25                                   & 56.39                                   & 64.11                                 & 74.06                                 & \underline{90.13}                     & 71.17                                 & 70.21                                   \\
                                                                          & BN                                 & \underline{54.83}                        & \underline{62.94}                     & \underline{73.81}                     & \underline{86.93}                     & \underline{69.63}                       & \underline{56.64}                       & \underline{64.19}                     & \underline{74.88}                     & 90.01                                 & \underline{71.43}                     & \underline{70.53}                       \\
                                                                          & Ours                               & \textbf{61.78}                           & \textbf{69.96}                        & \textbf{75.57}                        & \textbf{89.32}                        & \textbf{74.16}                          & \textbf{64.74}                          & \textbf{71.70}                        & \textbf{75.99}                        & \textbf{91.61}                        & \textbf{76.01}                        & \textbf{75.08}                          \\
                \midrule

\multirow{3}{*}{GPT-4o mini}                    & DP                                 & 45.73                                    & 56.02                                 & 66.31                                 & \underline{84.86}                     & 63.23                                   & 43.97                                   & 53.05                                 & 67.92                                 & 85.73                                 & 62.67                                 & 62.95                                   \\
                                                                          & BN                                 & \underline{46.09}                        & \underline{56.69}                     & \underline{66.89}                     & \underline{84.86}                     & \underline{63.63}                       & \underline{44.16}                       & \underline{53.45}                     & \underline{68.27}                     & \underline{86.50}                     & \underline{63.10}                     & \underline{63.36}                       \\
                                                                          & Ours                               & \textbf{47.63}                           & \textbf{58.99}                        & \textbf{72.41}                        & \textbf{90.09}                        & \textbf{67.28}                          & \textbf{46.03}                          & \textbf{57.00}                        & \textbf{74.57}                        & \textbf{90.74}                        & \textbf{67.09}                        & \textbf{67.18}                          \\
                \midrule

\multirow{3}{*}{gpt-oss-20b}                    & DP                                 & 55.04                                    & 65.47                                 & 80.05                                 & 81.41                                 & 70.49                                   & 53.05                                   & 60.50                                 & 80.88                                 & \underline{81.97}                     & 69.10                                 & 69.80                                   \\
                                                                          & BN                                 & \underline{55.74}                        & \underline{65.89}                     & \underline{80.65}                     & \underline{81.99}                     & \underline{71.07}                       & \underline{53.83}                       & \underline{60.59}                     & \underline{81.37}                     & \underline{81.97}                     & \underline{69.44}                     & \underline{70.25}                       \\
                                                                          & Ours                               & \textbf{58.21}                           & \textbf{66.75}                        & \cellcolor{secondcolor}\textbf{85.33} & \textbf{89.20}                        & \textbf{74.87}                          & \textbf{59.60}                          & \textbf{66.88}                        & \cellcolor{thirdcolor}\textbf{87.29}  & \textbf{88.81}                        & \textbf{75.65}                        & \textbf{75.26}                          \\
                \midrule

\multirow{3}{*}{Claude 3 Haiku}                 & DP                                 & 46.12                                    & 59.28                                 & 79.84                                 & 78.35                                 & 65.90                                   & 45.97                                   & 58.64                                 & 80.37                                 & 82.90                                 & 66.97                                 & 66.43                                   \\
                                                                          & BN                                 & \underline{46.75}                        & \underline{59.46}                     & \underline{80.12}                     & \underline{79.40}                     & \underline{66.43}                       & \underline{46.35}                       & \underline{58.97}                     & \underline{81.36}                     & \underline{82.97}                     & \underline{67.41}                     & \underline{66.92}                       \\
                                                                          & Ours                               & \textbf{52.33}                           & \textbf{66.74}                        & \cellcolor{thirdcolor}\textbf{84.30}  & \textbf{87.37}                        & \textbf{72.69}                          & \textbf{50.89}                          & \textbf{64.35}                        & \textbf{84.33}                        & \textbf{90.11}                        & \textbf{72.42}                        & \textbf{72.55}                          \\
                \midrule

\multirow{3}{*}{Mistral Small 3.2}              & DP                                 & 47.02                                    & 58.20                                 & 71.11                                 & 87.87                                 & 66.05                                   & \underline{48.58}                       & \underline{59.91}                     & \underline{74.34}                     & 86.18                                 & \underline{67.25}                     & 66.65                                   \\
                                                                          & BN                                 & \underline{48.10}                        & \underline{58.41}                     & \underline{71.52}                     & \underline{88.11}                     & \underline{66.54}                       & 48.35                                   & 59.57                                 & \underline{74.34}                     & \underline{86.32}                     & 67.15                                 & \underline{66.84}                       \\
                                                                          & Ours                               & \textbf{54.27}                           & \textbf{64.13}                        & \textbf{76.59}                        & \textbf{90.13}                        & \textbf{71.28}                          & \textbf{55.12}                          & \textbf{63.00}                        & \textbf{83.95}                        & \textbf{91.47}                        & \textbf{73.39}                        & \textbf{72.33}                          \\
                \midrule

\multirow{6}{*}{Llama 3.3 70B}                  & DP                                 & 44.26                                    & 56.02                                 & 66.89                                 & 85.48                                 & 63.16                                   & 38.16                                   & 50.94                                 & 70.74                                 & 84.30                                 & 61.04                                 & 62.10                                   \\
                                                                          & BN                                 & 45.15                                    & 56.49                                 & 66.97                                 & 85.87                                 & 63.62                                   & 40.24                                   & 51.16                                 & 71.50                                 & 84.55                                 & 61.86                                 & 62.74                                   \\
                                                                          & CoT                                & 45.39                                    & 57.10                                 & 67.21                                 & 85.83                                 & 63.88                                   & 39.66                                   & 51.19                                 & 71.72                                 & 84.93                                 & 61.88                                 & 62.88                                   \\
                                                                          & ReConcile                          & \underline{48.28}                        & 57.18                                 & 69.74                                 & \underline{88.91}                     & 66.03                                   & 45.45                                   & \underline{57.21}                     & \underline{75.89}                     & \underline{87.34}                     & 66.47                                 & 66.25                                   \\
                                                                          & DyLAN                              & 47.64                                    & \underline{58.19}                     & \underline{71.76}                     & 87.97                                 & \underline{66.39}                       & \underline{46.86}                       & 56.70                                 & 75.24                                 & 87.32                                 & \underline{66.53}                     & \underline{66.46}                       \\
                                                                          & Ours                               & \textbf{52.50}                           & \textbf{63.81}                        & \textbf{75.79}                        & \textbf{90.65}                        & \textbf{70.69}                          & \textbf{50.05}                          & \textbf{60.93}                        & \textbf{78.94}                        & \textbf{90.89}                        & \textbf{70.20}                        & \textbf{70.45}                          \\
                \midrule

\multirow{6}{*}{Gemini 3 Flash Preview}         & DP                                 & 54.09                                    & 63.35                                 & 64.22                                 & 89.59                                 & 67.81                                   & 57.12                                   & 65.89                                 & 68.67                                 & 90.63                                 & 70.58                                 & 69.20                                   \\
                                                                          & BN                                 & 54.70                                    & 63.48                                 & 64.81                                 & 89.95                                 & 68.24                                   & 57.19                                   & 66.23                                 & 69.21                                 & 90.79                                 & 70.86                                 & 69.55                                   \\
                                                                          & CoT                                & 55.11                                    & 63.64                                 & 65.70                                 & 89.72                                 & 68.54                                   & 57.64                                   & 66.68                                 & 69.23                                 & 91.24                                 & 71.20                                 & 69.87                                   \\
                                                                          & ReConcile                          & \underline{58.62}                        & \underline{67.91}                     & 74.89                                 & \underline{91.33}                     & 73.19                                   & \underline{61.22}                       & 68.50                                 & \underline{76.08}                     & 90.82                                 & \underline{74.16}                     & \underline{73.67}                       \\
                                                                          & DyLAN                              & 57.29                                    & 67.37                                 & \underline{77.54}                     & 90.94                                 & \underline{73.29}                       & 59.60                                   & \underline{68.81}                     & 74.49                                 & \underline{91.28}                     & 73.55                                 & 73.42                                   \\
                                                                          & Ours                               & \textbf{60.47}                           & \cellcolor{thirdcolor}\textbf{70.27}  & \textbf{81.73}                        & \cellcolor{bestcolor}\textbf{93.07}   & \textbf{76.39}                          & \textbf{64.91}                          & \cellcolor{secondcolor}\textbf{72.34} & \textbf{81.88}                        & \cellcolor{secondcolor}\textbf{93.64} & \cellcolor{thirdcolor}\textbf{78.19}  & \textbf{77.29}                          \\
                \midrule

\multirow{6}{*}{Qwen3-235B-A22B}                & DP                                 & 57.65                                    & 67.30                                 & 74.90                                 & 86.20                                 & 71.51                                   & 60.35                                   & 66.62                                 & 79.66                                 & \underline{90.57}                     & 74.30                                 & 72.91                                   \\
                                                                          & BN                                 & 57.99                                    & 67.61                                 & 75.75                                 & 86.41                                 & 71.94                                   & 60.87                                   & 67.84                                 & 80.17                                 & 90.57                                 & 74.86                                 & 73.40                                   \\
                                                                          & CoT                                & 58.20                                    & 67.96                                 & 75.46                                 & 86.53                                 & 72.04                                   & 60.76                                   & 67.61                                 & 80.37                                 & 90.84                                 & 74.90                                 & 73.47                                   \\
                                                                          & ReConcile                          & 60.22                                    & 68.74                                 & 78.82                                 & \underline{89.97}                     & 74.44                                   & \underline{63.04}                       & 69.52                                 & 84.69                                 & \underline{91.79}                     & \underline{77.26}                     & 75.85                                   \\
                                                                          & DyLAN                              & \underline{61.42}                        & \underline{68.97}                     & \underline{80.06}                     & 87.89                                 & \underline{74.59}                       & 61.81                                   & \underline{70.53}                     & \underline{85.24}                     & 91.22                                 & 77.20                                 & \underline{75.89}                       \\
                                                                          & Ours                               & \textbf{63.13}                           & \cellcolor{secondcolor}\textbf{71.36} & \textbf{83.22}                        & \textbf{91.36}                        & \cellcolor{secondcolor}\textbf{77.27}   & \cellcolor{secondcolor}\textbf{66.46}   & \cellcolor{thirdcolor}\textbf{71.94}  & \cellcolor{bestcolor}\textbf{88.36}   & \cellcolor{bestcolor}\textbf{93.82}   & \cellcolor{bestcolor}\textbf{80.15}   & \cellcolor{secondcolor}\textbf{78.71}   \\
                \midrule

\multirow{6}{*}{DeepSeek-V3.2}                  & DP                                 & 57.25                                    & 66.21                                 & 74.52                                 & 86.14                                 & 71.03                                   & 59.85                                   & 66.95                                 & 77.59                                 & 88.42                                 & 73.20                                 & 72.12                                   \\
                                                                          & BN                                 & 57.93                                    & 66.43                                 & 74.79                                 & 87.02                                 & 71.54                                   & 60.47                                   & 67.38                                 & 77.72                                 & 88.92                                 & 73.62                                 & 72.58                                   \\
                                                                          & CoT                                & 58.46                                    & 66.87                                 & 75.68                                 & 86.71                                 & 71.93                                   & 60.83                                   & 67.53                                 & 77.86                                 & 88.08                                 & 73.58                                 & 72.75                                   \\
                                                                          & ReConcile                          & \cellcolor{thirdcolor}63.44              & 69.82                                 & 81.12                                 & 89.77                                 & 76.04                                   & \cellcolor{thirdcolor}\underline{65.98} & \underline{69.70}                     & 83.56                                 & \underline{90.54}                     & 77.45                                 & 76.74                                   \\
                                                                          & DyLAN                              & \cellcolor{secondcolor}\underline{64.37} & \underline{69.96}                     & \underline{82.20}                     & \underline{91.47}                     & \cellcolor{thirdcolor}\underline{77.00} & 64.55                                   & 69.41                                 & \underline{86.84}                     & 90.04                                 & \underline{77.71}                     & \cellcolor{thirdcolor}\underline{77.36} \\
                                                                          & Ours                               & \cellcolor{bestcolor}\textbf{66.36}      & \cellcolor{bestcolor}\textbf{72.21}   & \cellcolor{bestcolor}\textbf{85.64}   & \cellcolor{secondcolor}\textbf{92.35} & \cellcolor{bestcolor}\textbf{79.14}     & \cellcolor{bestcolor}\textbf{67.62}     & \cellcolor{bestcolor}\textbf{72.46}   & \cellcolor{secondcolor}\textbf{87.32} & \textbf{92.93}                        & \cellcolor{secondcolor}\textbf{80.08} & \cellcolor{bestcolor}\textbf{79.61}     \\
                \bottomrule
        \end{tabular}}
        \vspace{-0.5em}
        \caption{Performance comparison across different models and methods on
        two splits. \textbf{DP}: Direct Prompting, \textbf{BN}: Best-of-N, \textbf{Ours}:
        RoadMapper with DPO training on Qwen3-32B, \textbf{SS}: StepScore, \textbf{LS}:
        LogicScore, \textbf{DegS}: DegreeScore, \textbf{DepS}: DepthScore. The
        top two results are highlighted in \textbf{bold} and \underline{underlined},
        respectively.}
        \label{tab:performance_comparison}
\end{table*}
\subsection{Inference Procedure}
We present the inference procedure of RoadMapper in Table~\ref{tab:inference_procedure}:
When a complex research problem is inputted, the $\mathcal{I}$ agent first
generates an initial roadmap, which is then augmented with knowledge by the $\mathcal{K}$
agent. Subsequently, the system enters an iterative ``critique-revise-evaluate''
process, supported by agents $\mathcal{L}$, $\mathcal{G}$, and $\mathcal{R}$. This
process continues until the roadmap achieves the \emph{passing score} or reaches
the maximum number of iterations.
  \section{Experiments}
\label{sec:experiments}

We present the experimental setup, main results, and a series of comprehensive analyses
in this section, with more details shown in Appendix~\ref{app:details_experiments}.

\subsection{Experimental Setup}
\label{sec:experiment_setup}

\paragraph{Evaluation Metrics.}
We evaluate from structure (\ie{ \textbf{DegreeScore} and \textbf{DepthScore}}) and
content (\ie{ \textbf{StepScore} and \textbf{LogicScore}}). Since content metrics
are not amenable to rule-based calculation, we align with established Long-form
QA evaluation methods~\citep{tan-etal-2024-proxyqa,wang-etal-2024-leave} and employ
\texttt{GPT-4o mini} to score the generated roadmaps with reference to the
golden ones.

\begin{itemize}[leftmargin=*]
    \item \textbf{DegreeScore} represents the magnitude of out-degree
        differences between the output and the golden roadmap, computed by:
        \begin{equation}
            \resizebox{0.85\linewidth}{!}{$DegreeScore = \left(1 - \frac{|Deg_{g}-
            Deg_{o}|}{Deg_{g}}\right) \times 100$}.
        \end{equation}

        Here, $Deg_{g}$ and $Deg_{o}$ denote the out-degree of the golden
        roadmap and the output roadmap.

    \item \textbf{DepthScore} represents the magnitude of depth differences
        between the output and the golden roadmap, computed by:
        \begin{equation}
            \resizebox{0.85\linewidth}{!}{$DepthScore = \left(1 - \frac{|Dep_{g}-
            Dep_{o}|}{Dep_{g}}\right) \times 100$}.
        \end{equation}

        Here, $Dep_{g}$ and $Dep_{o}$ denote the depth of the golden roadmap and
        the output roadmap.

    \item \textbf{StepScore} represents the key step score of the roadmap, which
        reflects the effectiveness of the roadmap. A higher score indicates that
        it better embodies the key steps marked in the golden roadmap (introduced
        in Section~\ref{sec:expert_guided_generation}).

    \item \textbf{LogicScore} represents the internal logical score of the
        roadmap, which reflects the logical coherence of the roadmap. A higher
        score indicates that it aligns more closely with the research
        methodology of the golden roadmap.
\end{itemize}

\paragraph{Baselines.}
We compare RoadMapper with several baseline methods, including different prompting
strategies and multi-agent systems.

\begin{itemize}[leftmargin=*]
    \itemsep0em

    \item \textbf{Direct Prompting} prompts the model with the research problem
        and the roadmap generation task. In our implementation, it uses the same
        instruction template as the $\mathcal{I}$ agent.

    \item \textbf{Best-of-N} strategy executes Direct Prompting independently
        for $N$ times and selects the best output as the final result.

    \item \textbf{CoT}~\citep{NEURIPS2022_9d560961} prompting encourages the model
        to explicitly decompose the research problem into intermediate reasoning
        steps before producing the final roadmap.

    \item \textbf{ReConcile}~\citep{chen-etal-2024-reconcile} organizes multiple
        agents in a round-table discussion, conducting multiple rounds of
        deliberation where agents attempt to persuade each other to improve the
        roadmap and ultimately reach a consensus.

    \item \textbf{DyLAN}~\citep{liu2024dynamicllmpoweredagentnetwork} adopts a dynamic
        workflow paradigm, where a subset of agents is dynamically selected to
        participate in roadmap generation, rather than involving all agents.
\end{itemize}

\paragraph{Implementation Details.}
\textbf{(1)} We conduct experiments on 11 LLMs, including both open-source and proprietary
models of various sizes from different manufacturers; \textbf{(2)} The maximum number
of iterations for the ``critique-revise-evaluate'' process is set to 5; \textbf{(3)}
The \emph{passing score} of $\mathcal{E}$ agent is set to 80; \textbf{(4)} The
maximum number of skill points that $\mathcal{K}$ agent retrieves from Skill-Repo
is set to 30; \textbf{(5)} We use Qwen3 32B with DPO training as $\mathcal{E}$
agent.

\subsection{Main Results}
\label{sec:main_results_of_roadmapper}

Table~\ref{tab:performance_comparison} presents the main results of our experiments.
There are two main conclusions as follows:

\textit{\textbf{Prior methods cannot effectively address the task of generating
research roadmaps.}} Our experimental results reveal that direct prompting leads
to suboptimal performance across all metrics. For instance, Llama 3.3 70B attained
an average score of 61.04 and a notably low StepScore of 38.16 on the Chinese split.
Best-of-N and CoT improved performance by increasing computation, but the gains are
marginal and inconsistent. ReConcile and DyLAN achieved limited improvements via
multi-agent strategies. For example, the average improvement of Qwen3-235B-A22B
is only 2.98.

\textit{\textbf{RoadMapper significantly improves the performance of LLMs in
roadmap generation tasks.}} Data show that RoadMapper achieves SOTA performance across
all base models in English and Chinese splits. For instance, compared to Direct Prompting,
the average performance improvements of Llama 3.3 70B on English and Chinese
splits are \textbf{7.53} and \textbf{9.16}, respectively, and DeepSeek-V3.2 also
achieves average performance improvements of \textbf{8.11} and \textbf{6.88},
respectively. Moreover, compared with ReConcile, RoadMapper delivers a further stable
improvement of \textbf{4.20} for Llama 3.3 70B. This enhancement exhibits the same
trend across models and metrics, demonstrating the effectiveness, robustness,
and generalizability of RoadMapper.

\subsection{Ablation Study}
\begin{figure*}[t!]
    \centering
    \includegraphics[width=0.95\textwidth]{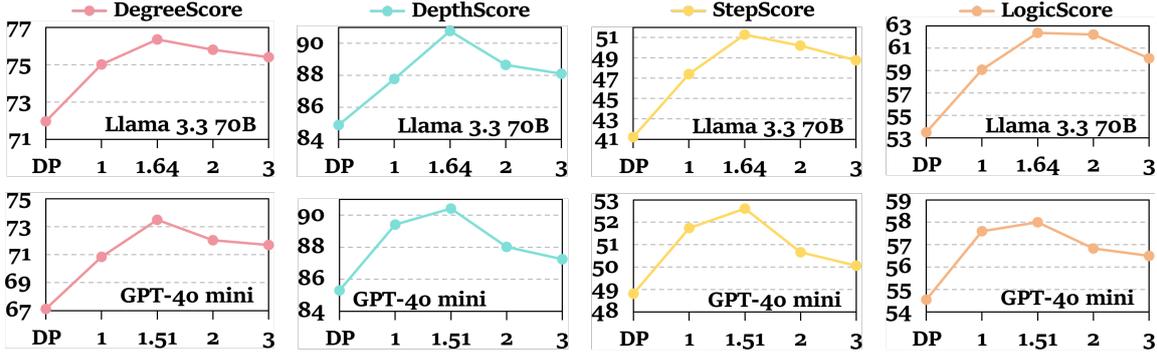}
    \caption{Quantitative analysis between enforced N iterations of ``critique-revise''
    and performance. \textbf{DP} denotes Direct Prompting. The X-axis represents
    the rounds of ``critique-revise'' and the Y-axis shows the evaluation metrics.}
    \label{fig:quantity_analysis}
\end{figure*}

We conduct an ablation study with results shown in Table~\ref{tab:method_comparison}.
There are two main conclusions:

\textbf{\textit{All of the agents $\mathcal{K}$, $\mathcal{L}$, and
$\mathcal{G}$ play positive roles.}} The data show that the complete RoadMapper outperforms
all variants across all metrics, and removing any single agent or merging agents
$\mathcal{L}$ and $\mathcal{G}$ results in varying degrees of performance
degradation. For instance, excluding $\mathcal{K}$ leads to a decline of 5.84 in
StepScore with an overall decrease of 3.72, while removing $\mathcal{L}$ leads to
a decline of 5.43 on LogicScore. Meanwhile, the performance degradation on SC variant
demonstrates the rationality of separating logic and granularity critiques.

\textbf{\textit{DPO training plays a positive role, depending on the backbone model
size.}} The data show that ablating DPO from agent $\mathcal{E}$ leads to an
average performance degradation of 2.21. Meanwhile, when using Qwen3-14B as the
backbone, the performance drops by 1.79. Notably, the Qwen3-8B backbone with DPO
performs worse than the w/o DPO variant, possibly due to its inherent lack of
capacity.

\begin{table}[t!]
    \centering
    \resizebox{\columnwidth}{!}{
    \begin{tabular}{lcccc|c}
        \toprule \textbf{Method}         & \textbf{SS}    & \textbf{LS}    & \textbf{DegS}  & \textbf{DepS}  & \textbf{Avg.}  \\
        \midrule w/o $\mathcal{K}$ agent & 45.44          & 59.60          & 73.11          & 88.76          & 66.73          \\
        w/o $\mathcal{L}$ agent          & 48.29          & 56.94          & 74.17          & 88.46          & 66.97          \\
        w/o $\mathcal{G}$ agent          & 47.51          & 58.74          & 74.80          & 87.37          & 67.11          \\
        w/o DPO                          & 49.83          & 60.20          & 74.67          & 88.25          & 68.24          \\
        DPO-8B                           & 49.27          & 58.47          & 73.02          & 87.62          & 67.10          \\
        DPO-14B                          & 49.77          & 60.69          & 75.30          & 88.87          & 68.66          \\
        SC                               & 48.52          & 58.98          & 74.29          & 87.21          & 67.25          \\
        \midrule \textbf{RoadMapper}     & \textbf{51.28} & \textbf{62.37} & \textbf{77.37} & \textbf{90.77} & \textbf{70.45} \\
        \bottomrule
    \end{tabular}
    }
    \caption{ Ablation study of RoadMapper. \textbf{w/o}: without, \textbf{Splits}:
    English \& Chinese, \textbf{Base Model}: Llama 3.3 70B, \textbf{DPO-8B}: use
    Qwen3-8B as backbone, \textbf{DPO-14B}: use Qwen3-14B as backbone, \textbf{SC}:
    merging the agents $\mathcal{L}$ and $\mathcal{G}$ into one agent for logic and
    granularity critique. }
    \label{tab:method_comparison}
\end{table}
\subsection{Efficiency and Experts Evaluation}
\paragraph{Early Stopping Efficiency.}
We analyzed the relationship between iterations of the ``critique-revise'' and performance,
and found that \textbf{\textit{RoadMapper achieves an early stopping mechanism
and thus balances performance and cost}}. As shown in Figure~\ref{fig:quantity_analysis},
RoadMapper averages 1.64 and 1.51 iterations for Llama 3.3 70B and GPT-4o mini respectively.
Compared to fewer iterations, increased iteration counts improve all metrics;
compared to excessive iterations, early stopping maintains high performance while
reducing computational costs, demonstrating the effectiveness of agent
$\mathcal{E}$.

\begin{figure}[htbp]
    \centering
    \includegraphics[width=\linewidth]{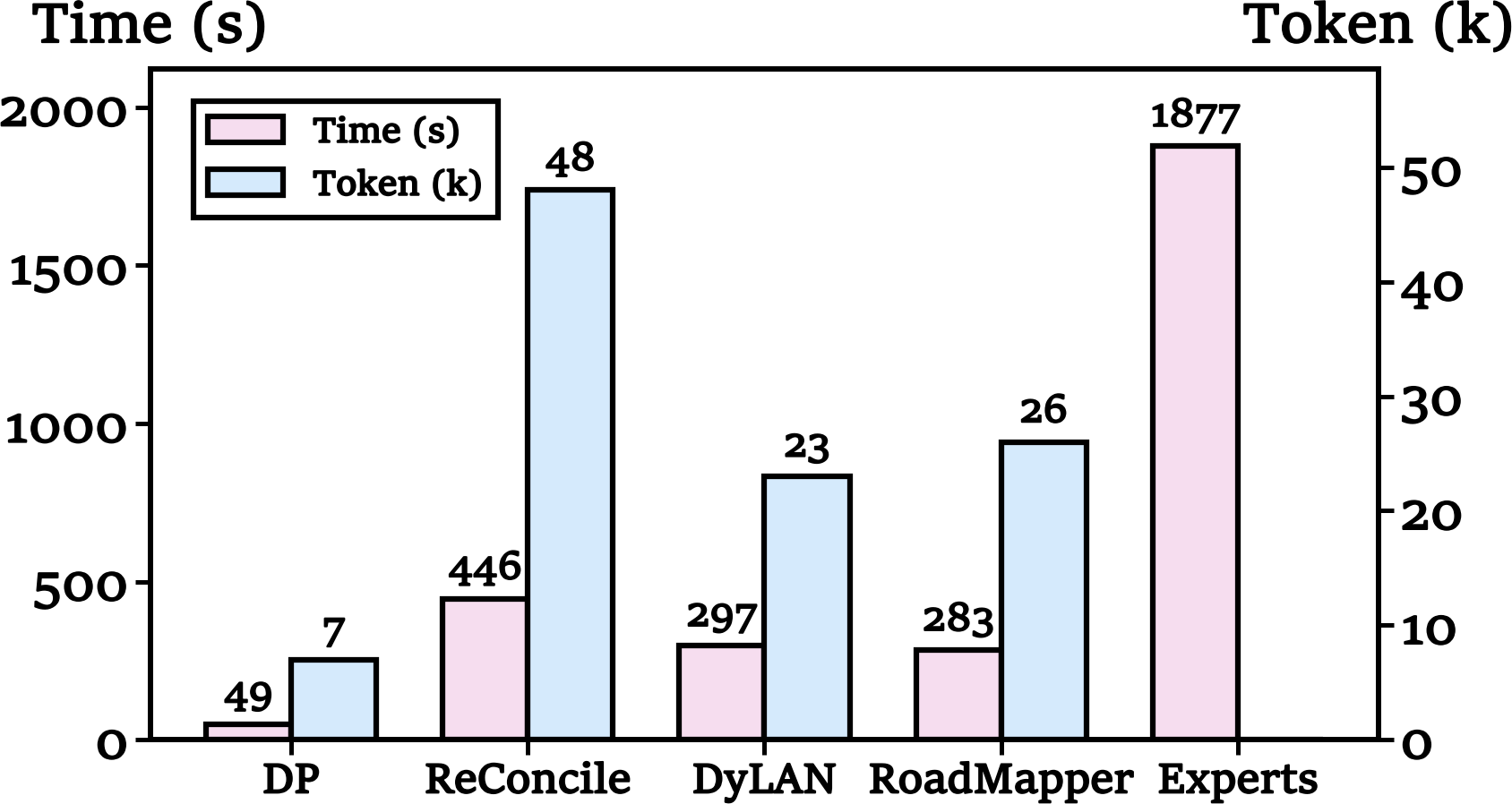}
    \caption{Cost comparison between methods on the roadmap generation task. \textbf{DP}
    denotes Direct Prompting.}
    \label{fig:cost_analysis}
\end{figure}
\paragraph{Time and Token Efficiency.}

As shown in Figure~\ref{fig:cost_analysis}, we comparatively analyzed the time and
token consumption of different methods in roadmap design. In terms of time, all LLM-based
automated methods significantly outperformed human experts, among which
RoadMapper requires the least time in complex systems, saving 36.5\% compared
with ReConcile. In terms of token consumption, RoadMapper saved 45.8\% compared with
ReConcile. These results highlight the substantial potential of RoadMapper for
practical applications.

\paragraph{Expert Evaluation.}

As shown in Appendix~\ref{sec:consistency_analysis}, we instruct seven experts to
evaluate the roadmaps across five dimensions: Logic Structure, Granularity,
Topic Relevance, Completeness, and Clarity, adopting the pairwise comparison paradigm.
The results show that (1) GPT-4o mini achieves a \textbf{93\% matching rate} with
experts evaluation, and (2) RoadMapper performs \textbf{better in 86\% of cases}.
These findings are consistent with the main results and validate the
effectiveness of RoadMapper.
  \section{Related Work}

\paragraph{Structured Content Generation.}
Structured content such as tables can help people quickly understand and
memorize knowledge. For instance,~\citet{jain-etal-2024-structsum} generate table
and mindmap summaries via specialized prompting;~\citet{li-etal-2023-sequence-sequence}
construct text-to-table systems using coordinated text encoders and table generators;~\citet{ren-etal-2023-constructing}
create program diagrams by analyzing text dependency relationships;~\citet{li-etal-2025-deepsolution}
implement engineering solution design through tree search-based Bi-point thinking.
Nevertheless, no effective method exists for generating research roadmaps addressing
complex research problems.

\paragraph{Multi-Agent Systems.}
LLMs may fail to adhere to multiple requirements when performing content
generation. Researchers suggest composite systems like multi-agent systems~\citep{amayuelas-etal-2024-multiagent,ijcai2024p890}.
Currently, relevant studies have explored applications in complex reasoning
tasks: simulating game decision-making~\citep{xu2024language}, clinical assistance~\citep{lu-etal-2024-triageagent},
and chart code generation~\citep{li-etal-2025-metal}. In contrast, our work
investigates the application of multi-agent systems in the roadmap generation
task.
  \section{Conclusion}
\label{sec:conclusion}

We identify a contradiction between the importance of roadmaps and the lack of related
research, and accordingly propose \textbf{RoadMap}, which evaluates the
capabilities of LLMs in research roadmap generation tasks. Based on the evaluation
results, we recognize three limitations of LLMs and propose \textbf{RoadMapper},
which accomplishes roadmap generation through coordinated work of multiple
agents. Experiments demonstrate that RoadMapper can significantly address the challenges
faced by LLMs and save much more time than experts require. We expect this work
to advance the research of LLMs in the \textbf{roadmap generation task}.
  \section*{Limitations}
First, RoadMapper relies on LLMs, which means that although we have employed the
currently best-performing prompts, the potential exists for more effective prompt
designs to further improve model performance. Second, while RoadMapper incorporates
an efficient early stopping mechanism, its computational cost remains relatively
higher than that of direct prompting, highlighting opportunities for future optimization.
Finally, due to limited computational resources, our experiments do not
encompass all available models, particularly those that are prohibitively expensive.
  \section*{Ethical Considerations}
The development of RoadMapper complies with the ACL ethics guidelines. This study
involved neither human subjects nor animal experimentation. All data were
sourced from open-source repositories in strict adherence to their respective
usage licenses, ensuring full privacy protection and the exclusion of any personally
identifiable information. To support both commercial and open-source
applications, RoadMapper is released under the Creative Commons Attribution 4.0 International
License (CC BY 4.0), while the associated codebase is distributed under the
Apache License 2.0. Furthermore, we have made consistent efforts to mitigate
potential bias and maintain absolute transparency throughout the dataset
construction and evaluation processes.

\section*{Use of AI Assistants}
This research was conducted with the primary intellectual contributions and core
scientific insights provided entirely by the authors. We acknowledge the use of AI-powered
tools such as Cursor to assist in data processing, code writing, and text polishing.
However, all key ideas, experimental design, analysis, and conclusions were
formulated through human-driven reasoning and expertise. The use of AI did not influence
the fundamental contributions or scientific integrity of this work.
  \section*{Acknowledgments}
This work is supported by the National Natural Science Foundation of China (Grant
Nos. 62473271, 62176026), the Fundamental Research Funds for the Beijing University
of Posts and Telecommunications (Grant No. 2025AI4S03), and the BUPT Innovation and
Entrepreneurship Support Program (Grant No. 2025-YC-A033). This work is also supported
by the Engineering Research Center of Information Networks, Ministry of
Education, China. We would also like to thank the anonymous reviewers and area chairs for
constructive discussions and feedback.
  \bibliography{acl2026}
  \clearpage
\appendix
\section{Details of RoadMap Benchmark}
\label{app:details_roadmap}

\subsection{Distribution of Publication Years}
\begin{figure}[htb]
    \centering
    \includegraphics[width=\columnwidth]{
        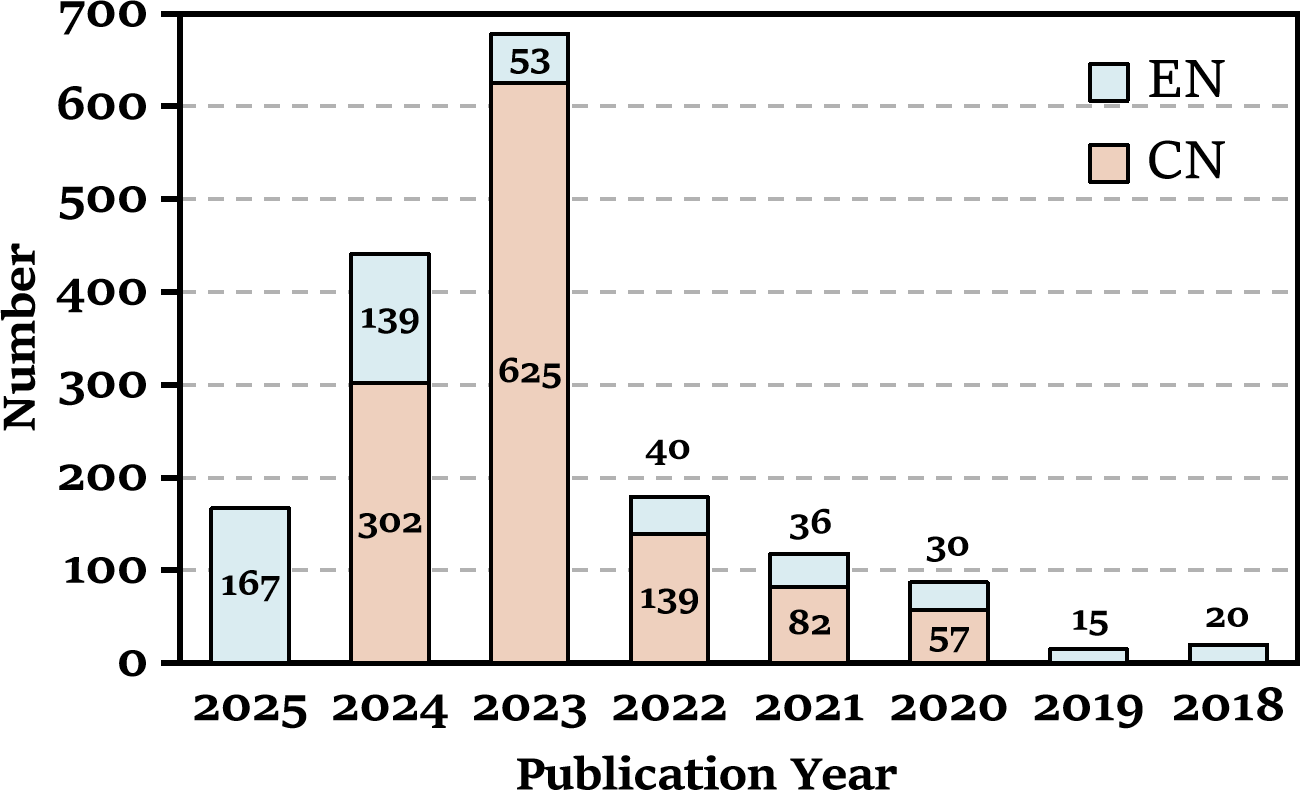
    }
    \caption{Distribution of the publication year of our collected dissertations.}
    \label{fig:publication_year_of_dissertations}
\end{figure}

As detailed in Section~\ref{data_collection}, we restricted our data collection
to dissertations published since 2018. Figure~\ref{fig:publication_year_of_dissertations}
presents the distribution of publication years for the entire dataset. Notably,
approximately 92.8\% of the dissertations were published within the last five years,
thereby ensuring the relevance and timeliness of the collected data.

\subsection{Parsing PDF Dissertations into Markdown Format}
We utilize MinerU\footnote{\url{https://mineru.net/}} to parse the PDF files of the
dissertations and extract Markdown results for subsequent processing. Here are
some details: (1) Our GPU specification is NVIDIA A40, each equipped with 48GB
VRAM; (2) We use version 2.0.6 of MinerU; (3) We employ \texttt{vlm-sglang-engine}
as the backend; (4) The total GPU time used for parsing is approximately 40 hours.

\subsection{Irrelevant Information Filtering in Markdown Files}
We use Python scripts to filter the parsed Markdown files and remove irrelevant information,
including author information, references, and invalid links. Key statistics regarding
the filtering process are shown in Table~\ref{tab:irrelevant_information_filtering}.
As described, this step removes approximately \textbf{37\%} of irrelevant
information from the dissertations, significantly reducing the computational cost
for subsequent processing.
\begin{table}[ht]
    \centering
    \resizebox{\columnwidth}{!}{{}{}
    \begin{tabular}{cccc} 
        \toprule \multicolumn{4}{c}{\textbf{Key Statistics Before and After Filtering}} \\ 
        \midrule \rowcolor{gray!10} \multicolumn{4}{c}{\textbf{English}}                \\ 
        \midrule                                                                       & \textbf{Total} & \textbf{Average} & \textbf{Reduction} \\ 
        \midrule Before                                                                & 210,390,046    & 420,780          & \errormark         \\
        After                                                                          & 131,579,990    & 263,159          & 37.46\%            \\
        \midrule \rowcolor{gray!10} \multicolumn{4}{c}{\textbf{Chinese}}                \\ 
        \midrule                                                                       & \textbf{Total} & \textbf{Average} & \textbf{Reduction} \\ 
        \midrule Before                                                                & 175,126,579    & 145,333          & \errormark         \\
        After                                                                          & 110,374,242    & 91,596           & 36.97\%            \\
        \bottomrule
    \end{tabular}}
    \caption{Key statistics of characters before and after filtering the parsed
    markdown files.}
    \label{tab:irrelevant_information_filtering}
\end{table}

\subsection{Template for Extraction}
As detailed in Section~\ref{sec:template_based_extraction}, we manually
formatted a template to extract core research problems and skill points from
collected dissertations, based on \texttt{Gemini 2.5 Flash}. Figure~\ref{fig:template_for_extraction}
illustrates the complete content of this template.

\subsection{Format Verification}
As stated in Section~\ref{sec:quality_assurance}, we have established strict
format requirements for the roadmap content. We developed a Python automated script
to assist in detecting any potential format errors, and any non-compliant nodes
were manually corrected by experts. We categorized all format errors into three types:
(1) \textbf{\textit{node format non-compliance}}, affecting \textbf{207} roadmaps;
(2) \textbf{\textit{level mismatch with index}}, affecting \textbf{83} roadmaps;
and (3) \textbf{\textit{incorrect index relationships between nodes}}, affecting
\textbf{193} roadmaps. Since a single roadmap may contain multiple types of
errors simultaneously, we ultimately performed format modifications on \textbf{329}
roadmaps, accounting for approximately \textbf{19.3\%} of the total.

\subsection{Redundancy Removal}
We implemented a comprehensive redundancy elimination pipeline based on semantic
embeddings to identify and remove potentially redundant skill points. Specifically,
each skill point was first encoded into an 8192-dimensional vector using the
\texttt{Qwen3-Embedding-8B} model. Subsequently, pairwise cosine similarities between
all skill points were computed using ChromaDB\footnote{\url{https://docs.trychroma.com/}}.
We heuristically flagged pairs with a cosine similarity exceeding $\tau = 0.6$ as
potentially redundant and submitted them to expert review for arbitration and merging.
  \section{Details of RoadMapper Methodology}
\label{app:details_roadmapper}

\subsection{\texorpdfstring{$\mathcal{K}$}{K} Agent}

The $\mathcal{K}$ agent is responsible for knowledge augmentation of the initial
roadmap by integrating both internal and external knowledge, operating in two sequential
stages.

In the first stage, it generates internal knowledge using the prompt illustrated
in Figure~\ref{fig:prompt_of_K}, while simultaneously retrieving relevant external
skill points from the Skill-Repo. To enable retrieval, we first encode all skill
points in the Skill-Repo into 8192-dimensional vectors using the \texttt{Qwen3-Embedding-8B}
model, based on their associated problem descriptions and skill point names.
During runtime, RoadMapper applies the same embedding method to compute the vector
representation of the given complex research problem. We then employ the
ChromaDB to retrieve the top-$K$ ($K$ = 30 in our experiments) most similar
skill points according to cosine similarity, which constitute the external knowledge
input for the $\mathcal{K}$ agent.

In the second stage, the agent performs knowledge augmentation operations by
leveraging both the generated internal knowledge and the retrieved external
knowledge, guided by the prompt shown in Figure~\ref{fig:prompt_of_K_for_augmentation}.

\subsection{\texorpdfstring{$\mathcal{E}$}{E} Agent}

We require the $\mathcal{E}$ agent to objectively evaluate the quality of the
roadmap along four dimensions (\ie{ Logic Structure, Granularity Degree, Topic Relevance, and Completeness})
and output the evaluation scores and detailed analysis, with each evaluation
score ranging from 0 to 100. The roadmap is only permitted to be output when the
evaluation score exceeds a predefined \emph{passing score} or when the maximum number
of iterations is reached. We present the prompt of the $\mathcal{E}$ agent in
Figure~\ref{fig:prompt_of_E}.

\subsection{Details of DPO Training}
We present key details of DPO training on $\mathcal{E}$ agent as follows:
\paragraph{Hardware and Software Environment.}

\begin{itemize}[leftmargin=*]
    \itemsep0em

    \item \textbf{GPU}: 4\,\texttimes\,NVIDIA RTX PRO 6000, each with 96 GB VRAM

    \item \textbf{CPU}: Intel Xeon Platinum 8470Q, 88 vCPUs

    \item \textbf{System memory}: 440 GB

    \item \textbf{OS}: Ubuntu 22.04

    \item \textbf{CUDA version}: 12.8

    \item \textbf{Framework}: PyTorch 2.8.0 + Python 3.12 + LlamaFactory 0.9.4.dev0
        + vLLM 0.13.0
\end{itemize}

\paragraph{Hyperparameters.}

\begin{itemize}[leftmargin=*]
    \itemsep0em

    \item \textbf{\texttt{pref\_beta}}: 0.2

    \item \textbf{\texttt{pref\_loss}}: \texttt{sigmoid}

    \item \textbf{Maximum sequence length}: 4,096 (Qwen3-32B), 8,192 (Qwen3-8B/14B)

    \item \textbf{Global batch size}: 16

    \item \textbf{Epochs}: 2, with 10\% warmup steps

    \item \textbf{Optimizer}: AdamW with learning rate = $5 \times 10^{-5}$,
        cosine decay, and weight decay

    \item \textbf{Precision}: BF16 mixed-precision training
\end{itemize}

\paragraph{Details of Constructing Preference Dataset.}
As described in Section~\ref{sec:dpo_training}, we prompt Qwen3-32B to evaluate roadmaps
collected from RoadMap and its construction pipeline, and these evaluations
constitute our candidate set. From this set, we select the candidates with the
highest and second-highest vote counts through expert voting to form the
positive and negative samples in the preference dataset, respectively. Meanwhile,
we carefully observed that the model may generate evaluations that do not conform
to the required format. To address this issue, we collected 36 model responses containing
formatting errors. For each erroneous response, we called domain experts to manually
correct the formatting. For every such error-correction pair, we constructed a preference
pair by treating the corrected evaluation as the positive sample and the original
malformed response as the negative sample, which were then incorporated into the
preference dataset for training. Experiments show that after augmenting the training
data with these format-oriented preference pairs, the model's formatting error
rate \textbf{drops significantly from 8\% to 3\%}, demonstrating the
effectiveness of our approach. Finally, our preference dataset comprises \textbf{818}
preference pairs, with \textbf{36 format-related pairs}.

\paragraph{Training time.}
The required GPU hours for training the Qwen3-8B/14B/32B models are
approximately 4/9/14, respectively.
  \section{Details of Experiments}
\label{app:details_experiments}
\subsection{Experimental Environment}

\paragraph{Base Environments.}
All our base experiments are conducted locally with the following details:

\begin{itemize}[leftmargin=*]
    \itemsep0em

    \item \textbf{OS}: Ubuntu 22.10

    \item \textbf{CPU}: Dual-socket Intel Xeon Gold 6148 (2.40 GHz), 40 cores
        per socket, 80 threads total

    \item \textbf{GPU}: 8\,\texttimes\,NVIDIA A40, each with 48 GB VRAM

    \item \textbf{GPU Driver}: 575.57.08

    \item \textbf{CUDA Version}: 11.8

    \item \textbf{cuDNN Version}: 8.x (compiled with CUDA 11.8)
\end{itemize}

\paragraph{Model Invocations.}
All model invocations are conducted via the OpenRouter\footnote{\url{https://openrouter.ai/}}
APIs with:

\begin{itemize}[leftmargin=*]
    \itemsep0em

    \item Request frequency: 200 rpm

    \item Max retries: 10 times/item

    \item Temperature: 1.0 (Inference), 0.2 (Evaluation)
\end{itemize}

\subsection{Base Models}
Table~\ref{tab:base_models} shows the list of base models used in the experiments,
including both open-source and proprietary models.
\begin{table}[t]
    \centering
    \resizebox{\columnwidth}{!}{
    \begin{tabular}{lcc} 
        \toprule \textbf{Model}                                           & \textbf{Institution} & \textbf{Model Card}                                    \\
        \midrule                                                           
        \rowcolor{gray!10} \multicolumn{3}{c}{\textbf{Open-Source Models}} \\ 
        \midrule Llama 3.1 8B                                             & Meta                 & ~~\citep{llama3_1}                                     \\
        Llama 3.3 70B                                                     & Meta                 & ~~\citep{llama3_3}                                     \\
        Llama 4 Maverick                                                  & Meta                 & ~~\citep{llama4}                                       \\
        Qwen3-14B                                                         & Alibaba              & ~~\citep{qwen3blog}                                    \\
        Qwen3-235B-A22B                                                   & Alibaba              & ~~\citep{qwen3blog}                                    \\
        gpt-oss-20b                                                       & OpenAI               & ~~\citep{openai2025gptoss120bgptoss20bmodel}           \\
        DeepSeek-V3.2                                                     & DeepSeek             & ~~\citep{deepseekai2025deepseekv32pushingfrontieropen} \\
        \midrule                                                           
        \rowcolor{gray!10} \multicolumn{3}{c}{\textbf{Proprietary Models}} \\ 
        \midrule GPT-4o mini                                              & OpenAI               & ~~\citep{openai2024gpt4ocard}                          \\
        Gemini 3 Flash Preview                                            & Google               & ~~\citep{google2025gemini3}                            \\
        Claude 3 Haiku                                                    & Anthropic            & ~~\citep{claude3}                                      \\
        Mistral Small 3.2                                                 & Mistral AI           & ~~\citep{mistral2025small32}                           \\
        \bottomrule
    \end{tabular}}
    \caption{Base models evaluated in main experiments.}
    \label{tab:base_models}
\end{table}

\subsection{Adaptability to Model Scale Variations}
RoadMapper can significantly improve LLM performance across various model sizes,
often enabling smaller models to outperform larger proprietary ones using conventional
methods. Our findings highlight that RoadMapper's enhancements are not confined
to large or top-tier models but extend effectively to smaller LLMs. For instance,
Llama 3.1 8B with RoadMapper achieves an average score of \textbf{67.82},
marking a substantial 7.20 improvement over its Direct Prompting baseline (60.62).
This enhanced performance notably surpasses the 66.43 average score of Claude 3 Haiku
using Direct Prompting. This demonstrates that RoadMapper effectively
democratizes high-quality roadmap generation, making advanced capabilities accessible
to models with more limited capacities and showcasing its \textbf{efficacy and
broad applicability regardless of model scale}.
  \section{Other Important Details}
\label{app:details_other}

\subsection{Comparison of Evaluation Prompts in Different Formats}
We observe that when employing GPT-4o mini as an evaluator to evaluate the content
quality of roadmaps, different prompt formats significantly influence the model's
performance. To quantitatively analyze this phenomenon and determine the optimal
prompting strategy, we conduct a comparative experiment. Specifically, we first
randomly select 100 roadmaps from the outputs generated by Llama 3.3 70B. Then,
we employ GPT-4o mini to evaluate these roadmaps using different prompting
strategies. Finally, we assess the performance of different prompting strategies
using two metrics:

\begin{itemize}[leftmargin=*]
    \itemsep0em

    \item \textbf{Consistency with Experts (CE)} measures the alignment between
        model evaluations and expert judgments. Specifically, we invite seven
        domain experts to vote for the best evaluation among all evaluations for
        each roadmap, and then calculate the vote rate for each prompting strategy.

    \item \textbf{Format Correctness (FC)} quantifies the reliability of
        structured output generation. We compute the probability that the model outputs
        conform to the required format across prompting strategies.
\end{itemize}

We test four prompting strategies, where each strategy's output format
requirements include both JSON-style and XML-style formats:

\begin{itemize}[leftmargin=*]
    \itemsep0em

    \item \textbf{AIH (All-In-Head)}: Both the reference roadmap and the evaluated
        roadmap are placed at the head of the prompt.

    \item \textbf{AIE (All-In-End)}: Both the reference roadmap and the evaluated
        roadmap are placed at the end of the prompt.

    \item \textbf{RHEE (Reference-Head-Evaluated-End)}: The reference roadmap is
        placed at the head of the prompt, while the evaluated roadmap is placed
        at the end of the prompt.

    \item \textbf{REEH (Reference-End-Evaluated-Head)}: The reference roadmap is
        placed at the end of the prompt, while the evaluated roadmap is placed
        at the head of the prompt.
\end{itemize}

As described in Table~\ref{tab:prompt_comparison}, the AIH strategy achieves the
best score on the CE metric. Moreover, the XML-style output format requirement
obtains higher scores on the FC metric. Therefore, we adopt the AIH strategy as
the prompting strategy for GPT-4o mini in our experiments, and require the model
to output evaluation in XML format.

\begin{table}[t]
    \centering
    \resizebox{0.7\columnwidth}{!}{
    \begin{tabular}{l|c|cc}
        \toprule \multirow{2.5}{*}{\textbf{Strategy}} & \textbf{CE (\%)} & \multicolumn{2}{c}{\textbf{FC (\%)}} \\
        \cmidrule(lr){2-2} \cmidrule(lr){3-4}         &                  & \textbf{JSON}                       & \textbf{XML} \\
        \midrule AIH                                  & 34               & 86                                  & 91           \\
        AIE                                           & 30               & 87                                  & 89           \\
        RHEE                                          & 17               & 84                                  & 83           \\
        REEH                                          & 19               & 82                                  & 83           \\
        \bottomrule
    \end{tabular}
    }
    \caption{ Comparison of different prompting strategies for GPT-4o mini evaluation.
    \textbf{CE}: Consistency with Experts, \textbf{FC}: Format Correctness. }
    \label{tab:prompt_comparison}
\end{table}

\subsection{Consistency Analysis Between Model Evaluation and Expert Judgments}
\label{sec:consistency_analysis}

To evaluate the quality of roadmaps generated by different methods from the perspective
of human experts and validate the reliability of using GPT-4o mini as an automated
evaluator, we conduct a consistency analysis by adopting a pairwise comparison
paradigm that aligns better with human cognitive processes~\citep{zheng2023judgingllmasajudgemtbenchchatbot, engelmann-etal-2024-arts, haak2025pairwisecomparisonbiasidentification}.

Specifically, we first randomly collected $N_{P}= 100$ research problems from
RoadMap and their corresponding roadmaps generated by both the Direct Prompting
and RoadMapper methods, using \texttt{DeepSeek-V3.2} as the base model. Each research
problem was paired with its respective roadmaps, forming $N_{P}$ evaluation
instances. Each evaluation instance is defined as a triplet:
\begin{equation}
    E_{i}= (P_{i}, \{R_{D_i}, R_{M_i}\}).
\end{equation}

Here, $P_{i}$ denotes the $i$-th research problem, for $i \in \{1, 2, \ldots, N_{P}
\}$, $R_{D_i}$ denotes the roadmap generated by the \textbf{Direct Prompting} method
for the research problem $P_{i}$, $R_{M_i}$ denotes the roadmap generated by the
\textbf{RoadMapper} method for the research problem $P_{i}$. It is important to
note that the order of $R_{D_i}$ and $R_{M_i}$ within the set
$\{ R_{D_i}, R_{M_i}\}$ is randomized to ensure anonymity during the evaluation
process.

During the human evaluation phase, we instructed $N_{E}= 7$ domain experts to serve
as human evaluators. For each evaluation instance $E_{i}$, every expert was prompted
to vote for the higher-quality roadmap across five dimensions: \textbf{Logic
Structure}, \textbf{Granularity Degree}, \textbf{Topic Relevance}, \textbf{Completeness},
and \textbf{Clarity}, denoted collectively as the dimension set
$\mathcal{D}= \{\text{LS}, \text{GD}, \text{TR}, \text{Co}, \text{Cl}\}$. The total
vote count for each roadmap is computed as:
\begin{align}
    V(R_{D_i}) & = \sum_{k=1}^{N_{E}}\sum_{j \in \mathcal{D}}v_{j}^{(k)}(R_{D_i}), \\
    V(R_{M_i}) & = \sum_{k=1}^{N_{E}}\sum_{j \in \mathcal{D}}v_{j}^{(k)}(R_{M_i}).
\end{align}

Here, $v_{j}^{(k)}(R_{D_i})$ and $v_{j}^{(k)}(R_{M_i})$ denote the binary vote
assigned by the $k$-th expert on dimension $j$, where $v_{j}^{(k)}(R_{D_i}) + v_{j}
^{(k)}(R_{M_i}) = 1$.

The final winner from experts is defined by:
\begin{equation}
    \resizebox{0.8\linewidth}{!}{ $ \displaystyle
    W_{i}^{\text{experts}}=
    \begin{cases}
        R_{D_i}, & \text{if }V(R_{D_i}) > V(R_{M_i}) \\
        R_{M_i}, & \text{otherwise}
    \end{cases} $ }.
\end{equation}

Here, $W_{i}^{\text{experts}}$ denotes the superior roadmap in instance $E_{i}$
as judged by the human experts.

On the model side, we derived the preference of GPT-4o mini by comparing the scalar
scores it assigned to $R_{D_i}$ and $R_{M_i}$, using the same evaluation
settings introduced in Section~\ref{sec:experiment_setup}. So, the final winner
from GPT-4o mini is defined by:
\begin{equation}
    \resizebox{0.8\linewidth}{!}{ $ \displaystyle
    W_{i}^{\text{evaluator}}=
    \begin{cases}
        R_{D_i}, & \text{if }S(R_{D_i}) > S(R_{M_i}) \\
        R_{M_i}, & \text{otherwise}
    \end{cases} $}.
\end{equation}

Here, $W_{i}^{\text{evaluator}}$ denotes the superior roadmap in instance $E_{i}$
as judged by GPT-4o mini as an automated evaluator.

Finally, we calculate the matching rate between $W_{i}^{\text{experts}}$ and
$W_{i}^{\text{evaluator}}$, described by:
\begin{equation}
    \text{MR}= \frac{\sum_{i=1}^{N_{P}}\mathbbm{1}[W_{i}^{\text{experts}}= W_{i}^{\text{evaluator}}]}{N_{P}}
    .
\end{equation}

Here, MR denotes the matching rate. The experimental results show that the MR between
the model evaluation and expert evaluation reaches 93\%, indicating that the scoring
of GPT-4o mini is highly consistent with human judgment, supporting its use as a
reliable evaluation proxy.

We further analyzed the distribution of expert evaluations, with results shown in
Table~\ref{tab:detail_distribution_of_human_experts}. The roadmaps generated by
RoadMapper were judged to be superior to those produced by Direct Prompting in
\textbf{86\%} of cases (\ie{ $\sum_{i=1}^{N_P}\mathbbm{1}[W_{i}^{\text{experts}}= R_{M_i}] = 86$}). Meanwhile,
RoadMapper shows significant advantages over Direct Prompting on three fine-grained
dimensions: \textbf{LS}, \textbf{GD}, and \textbf{Co}. These findings are highly
consistent with the main experimental results (shown in Section~\ref{sec:main_results_of_roadmapper})
and further validate the effectiveness of the RoadMapper method in the research
roadmap generation task.

\begin{table}[t]
    \centering
    \resizebox{\columnwidth}{!}{
    \begin{tabular}{c|ccccc|c}
        \toprule \textbf{Metric} & \textbf{LS} & \textbf{GD} & \textbf{TR} & \textbf{Co} & \textbf{Cl} & \textbf{Overall} \\
        \midrule Outcome         & 80          & 78          & 57          & 85          & 64          & 86               \\
        \bottomrule
    \end{tabular}
    }
    \caption{ Detailed human evaluation results across different dimensions. }
    \label{tab:detail_distribution_of_human_experts}
\end{table}

\subsection{Cross-Model Validation of Evaluators}

To validate potential bias in GPT-4o mini as an automatic evaluator, we introduce
Gemini 2.5 Flash and Qwen2.5 72B as alternative evaluators to conduct cross-model
consistency validation against GPT-4o mini. The experiment is based on the same set
of 100 research problem instances used in Appendix~\ref{sec:consistency_analysis},
and comprises two aspects:

\begin{itemize}[leftmargin=*]
    \itemsep0em

    \item \textbf{Correlation Analysis}: We computed the Spearman correlation coefficient
        between the scores assigned by Gemini 2.5 Flash and Qwen2.5 72B to the
        100 roadmaps and those assigned by GPT-4o mini. Results show that the correlation
        coefficient between Gemini 2.5 Flash and GPT-4o mini is \textbf{0.812}, and
        that between Qwen2.5 72B and GPT-4o mini is \textbf{0.849}. All p-values
        are less than 0.001, indicating a high degree of alignment in scoring trends
        across models.

    \item \textbf{Match Consistency Analysis}: We followed the pairwise comparison
        paradigm from Section~\ref{sec:consistency_analysis}: for each pair of
        roadmaps generated by Direct Prompting and RoadMapper, we determined the
        superior one based on the scores from Gemini 2.5 Flash or Qwen2.5 72B, and
        then compared this judgment with that of GPT-4o mini. We calculated the
        proportion of cases where both models agreed. Results indicate that Gemini
        2.5 Flash matches GPT-4o mini in \textbf{90\%} of cases, and Qwen2.5 72B
        matches in \textbf{93\%} of cases, corroborating the robustness of the
        evaluation outcomes.
\end{itemize}

The cross-model validation results shown in Table~\ref{tab:cross_model_validation_bias}
demonstrate that despite architectural differences, leading large language
models exhibit high consistency in evaluating the quality of research roadmaps, ensuring
the reliability of GPT-4o mini as an automated evaluation agent.

\begin{table}[t]
    \centering
    \resizebox{\columnwidth}{!}{
    \begin{tabular}{lccc}
        \toprule \textbf{Model}   & \textbf{Spearman R} & \textbf{P-Value} & \textbf{Match Rate (\%)} \\
        \midrule Gemini 2.5 Flash & 0.812               & $< 0.001$        & 90                       \\
        Qwen2.5 72B               & 0.849               & $< 0.001$        & 93                       \\
        \bottomrule
    \end{tabular}
    }
    \caption{ Cross-model validation results against GPT-4o mini for bias assessment.
    }
    \label{tab:cross_model_validation_bias}
\end{table}


\subsection{Stability Analysis of Evaluation Results}

We introduce our evaluation metrics in Section~\ref{sec:experiment_setup}, as described, the content metrics (\ie{ \textbf{StepScore}, \textbf{LogicScore}}) are content-based metrics that rely on the judgment of GPT-4o mini. So, it is crucial to demonstrate the stability of these metrics to ensure the validity of our experimental findings.

To assess the stability of the evaluation outcome from GPT-4o mini, we conduct repeated
evaluations. Specifically, we still utilize GPT-4o mini as the evaluator to re-evaluate
the experimental results generated by the Llama 3.3 70B model using the RoadMapper
method, on both the English and Chinese splits. For these repeated evaluations,
we compute the Pearson Correlation Coefficient, Mean Absolute Error, and Root
Mean Squared Error between the scores from the two independent runs.

As shown in Table~\ref{tab:stability_llama70b}, it demonstrates high stability in
the evaluation process. The Pearson correlation coefficients, ranging from 0.82 to
0.87 across all metrics and splits, indicate strong positive correlations
between two independent runs. Furthermore, the low MAE values (all below 1.5 points)
and RMSE values (all below 3.6 points) confirm that the absolute differences
between repeated scores are minimal. Collectively, these findings validate the
reliability and robustness of using GPT-4o mini as an evaluator for our content
metrics.

\begin{table}[t]
    \centering
    \resizebox{\columnwidth}{!}{
    \begin{tabular}{lc|ccc}
        \toprule \textbf{Metric}                      & \textbf{Split} & \textbf{Pearson R} & \textbf{MAE} & \textbf{RMSE} \\
        \midrule \multirow{2}{*}{\textbf{StepScore}}  & English        & 0.87               & 0.93         & 3.00          \\
                                                      & Chinese        & 0.84               & 1.41         & 3.52          \\
        \midrule \multirow{2}{*}{\textbf{LogicScore}} & English        & 0.83               & 1.11         & 2.75          \\
                                                      & Chinese        & 0.82               & 1.26         & 3.27          \\
        \bottomrule
    \end{tabular}
    }
    \caption{Stability analysis of content evaluation metrics for Llama 3.3 70B.
    \textbf{Pearson R}: Pearson Correlation Coefficient, \textbf{MAE}: Mean
    Absolute Error, \textbf{RMSE}: Root Mean Squared Error.}
    \label{tab:stability_llama70b}
\end{table}

\subsection{Significance Analysis}
\label{sec:significance_analysis}

To evaluate whether the performance gains of RoadMapper over Direct Prompting
are statistically significant, we conduct paired two-sided $t$-tests on the outputs
of gpt-oss-20b on the English split.

As shown in Table~\ref{tab:ttest_results}, RoadMapper achieves significant gains
on all metrics with $p < 0.001$, and all 95\% confidence intervals exclude zero.
For instance, on \texttt{StepScore}, the mean improvement is 3.17 with a 95\% CI
of [2.37, 3.98], indicating that RoadMapper can generate roadmaps with better coverage
of essential research steps. These results confirm that the performance gains
are statistically robust across all evaluation dimensions.

\begin{table}[t]
    \centering
    \resizebox{\columnwidth}{!}{
    \begin{tabular}{lcccc}
        \toprule \textbf{Metric} & \textbf{t} & \textbf{P-Value} & \textbf{95\% CI} & \textbf{Significance} \\
        \midrule StepScore       & -7.7261    & $< 0.001$        & [2.37, 3.98]     & ***                   \\
        LogicScore               & -3.9899    & $< 0.001$        & [0.65, 1.92]     & ***                   \\
        DegreeScore              & -6.8415    & $< 0.001$        & [3.77, 6.81]     & ***                   \\
        DepthScore               & -11.6542   & $< 0.001$        & [6.48, 9.11]     & ***                   \\
        \bottomrule
    \end{tabular}
    }
    \caption{Results of paired two-sided $t$-tests comparing RoadMapper vs. Direct
    Prompting on the English split. Significance levels: *** $p < 0.001$, ** $p <
    0.01$, * $p < 0.05$.}
    \label{tab:ttest_results}
\end{table}

\subsection{Impact of DPO on General Knowledge Abilities}
\label{appendix:dpo_general_knowledge}

To investigate whether DPO adversely affects the model's general knowledge and
reasoning capabilities, we conduct a controlled evaluation across four established
benchmarks spanning mathematical reasoning, code generation, and step-by-step problem
solving, using the backbone model Qwen3-32B and the DPO-optimized model:

\begin{itemize}[leftmargin=*]
    \itemsep0em

    \item \textbf{GSM8K}~\citep{cobbe2021trainingverifierssolvemath}: A dataset
        of 8,792 grade-school-level math word problems requiring multi-step
        arithmetic reasoning.

    \item \textbf{MATH-500}~\citep{hendrycksmath2021}: A curated subset of 500
        challenging high-school math problems from the MATH dataset, covering algebra,
        geometry, combinatorics, and calculus.

    \item \textbf{MBPP}~\citep{austin2021programsynthesislargelanguage}: The
        Most Basic Python Problems benchmark with 427 entry-level coding tasks
        evaluated by exact function match (pass@1).

    \item \textbf{LiveBench}~\citep{white2025livebench}: A dynamic benchmark
        designed to test real-world generalization. We evaluate on the reasoning
        split.
\end{itemize}

As shown in Table~\ref{tab:dpo_robustness}, the model maintains strong
performance across all tasks after DPO. On GSM8K, there is a minor drop from
79.83\% to 78.11\%. Similarly, MATH-500 sees a slight decrease from 88.40\% to 87.80\%.
These small regressions may stem from subtle shifts in decoding behavior or over-normalization
of reasoning paths during preference learning. Notably, the model improves on MBPP
(from 69.56\% to 71.90\%) and LiveBench (from 83.50\% to 85.50\%), indicating
that \textbf{the well-designed preference data not only enforces compliance but
can also reinforce coherent and effective problem-solving strategies}.

\begin{table}
    \centering
    \resizebox{\columnwidth}{!}{
    \begin{tabular}{lcc|c}
        \toprule \textbf{Benchmark} & \textbf{Backbone (\%)} & \textbf{DPO (\%)} & \textbf{$\Delta$ (\%)} \\
        \midrule GSM8K              & 79.83                  & 78.11             & -1.72                  \\
        MATH-500                    & 88.40                  & 87.80             & -0.60                  \\
        MBPP                        & 69.56                  & 71.90             & +2.34                  \\
        LiveBench                   & 83.50                  & 85.50             & +2.00                  \\
        \bottomrule
    \end{tabular}
    }
    \caption{Performance comparison between the backbone model and the DPO-finetuned
    model on general knowledge and reasoning benchmarks. The $\Delta$ column shows
    the absolute difference (DPO $-$ Backbone), indicating high overall
    stability across all tasks.}
    \label{tab:dpo_robustness}
\end{table}

\subsection{Case Study}
\begin{figure*}[t!]
    \centering
    \includegraphics[width=\textwidth]{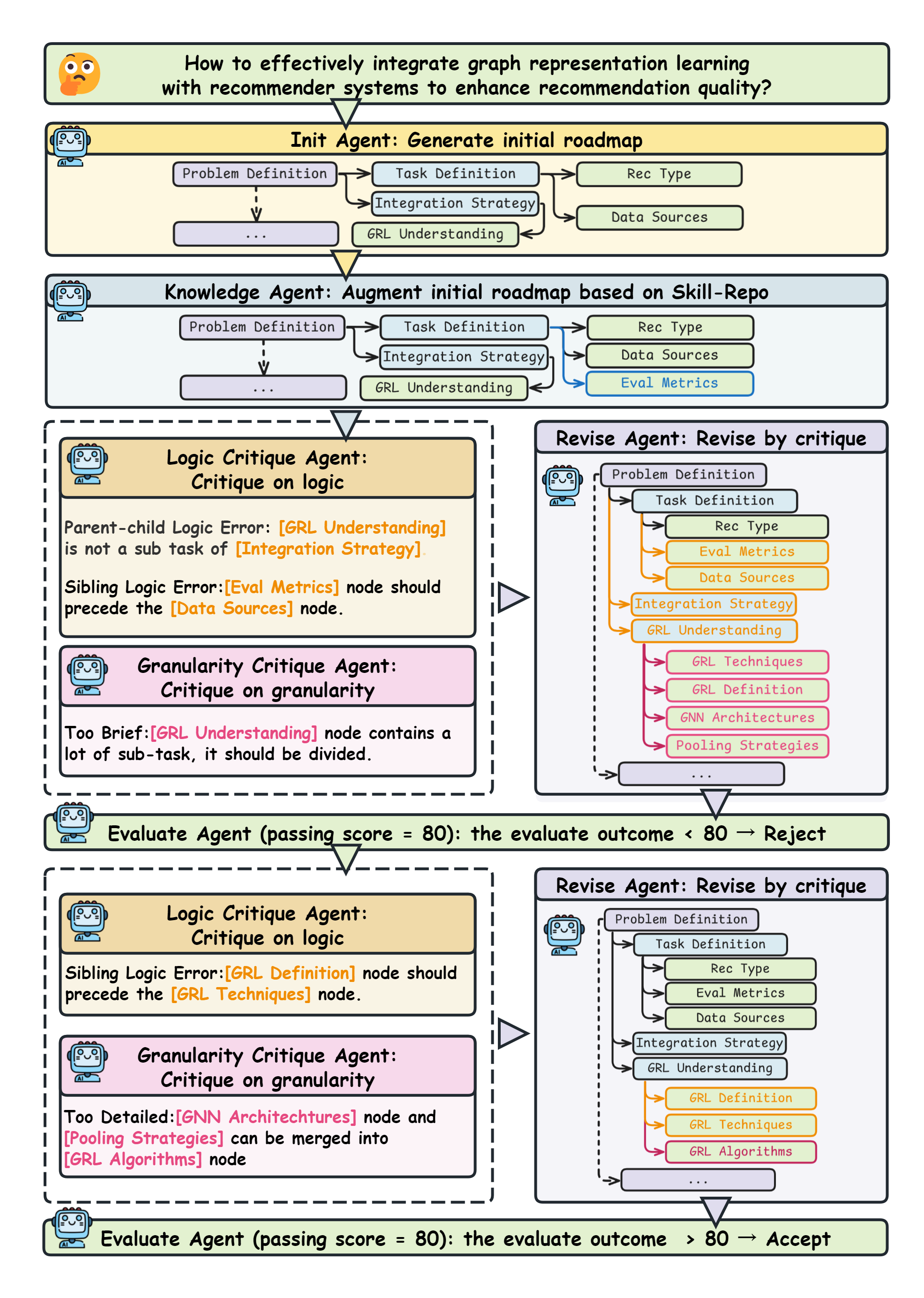}
    \caption{Case study of RoadMapper.}
    \label{fig:case_study}
\end{figure*}

Figure~\ref{fig:case_study} presents a simplified yet complete case study to
illustrate the operational process of RoadMapper. After the research problem is input,
the Init Agent generates an initial roadmap. Subsequently, the Knowledge Agent
inserts an \texttt{[Eval Metrics]} node. The process then enters the critique
phase, where the Revise Agent adjusts the node order and decomposes tasks according
to suggestions. The revised roadmap is evaluated by the Evaluator. Since it does
not meet the passing score, the output is rejected, and the process proceeds to the
next round of critique. The Revise Agent again adjusts the node order and merges
subtasks based on the improvement suggestions. The subsequently revised roadmap passes
the Evaluator's evaluation and is successfully output, ultimately yielding the
final roadmap. This case study perfectly demonstrates the initial generation,
knowledge augmentation, and iterative ``critique-revise-evaluate'' process of the
RoadMapper system.

\subsection{Example of Golden-Roadmap}
\begin{figure*}[t!]
    \centering
    \includegraphics[width=\textwidth]{
        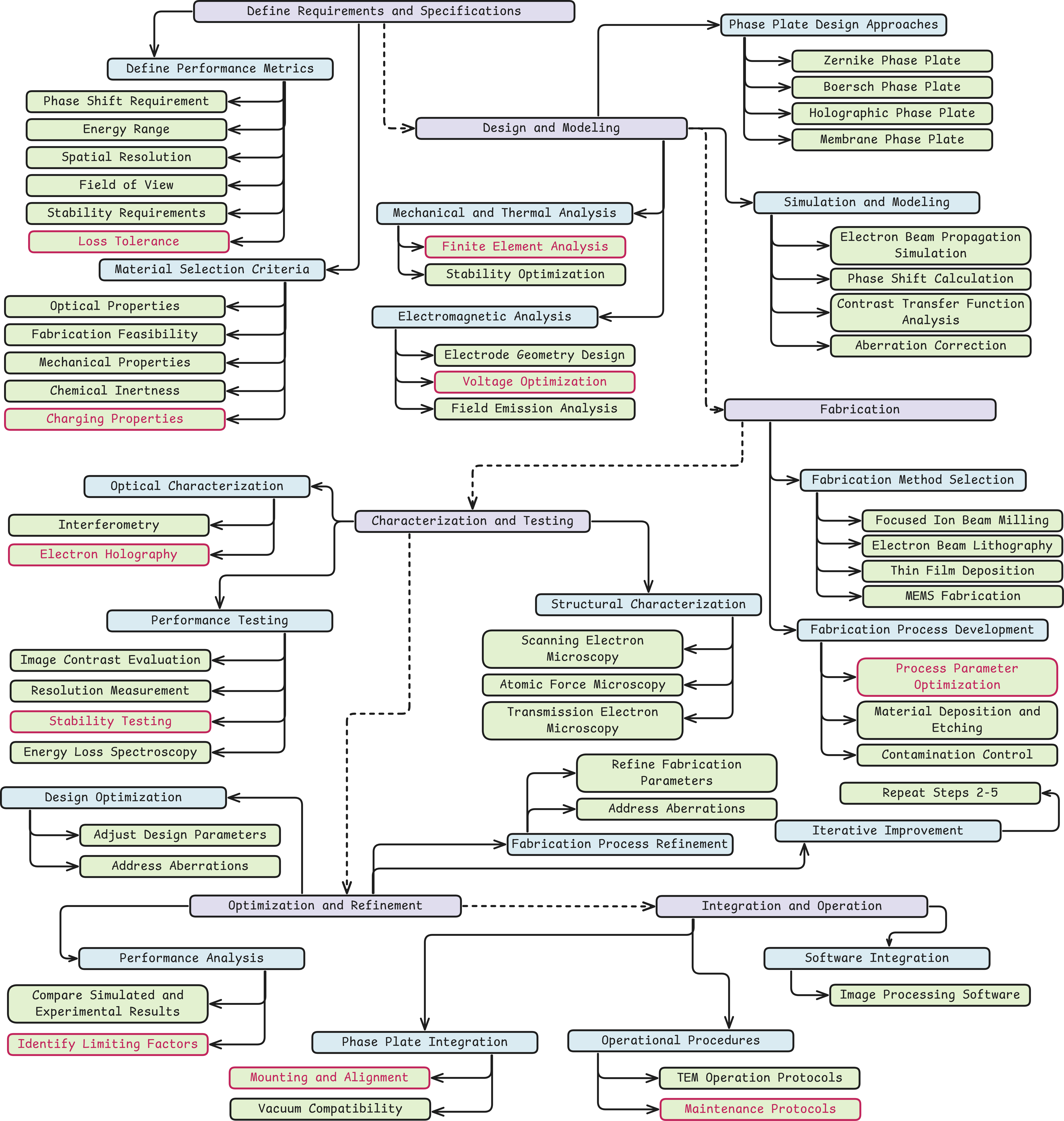
    }
    \caption{Example of Golden-Roadmap.}
    \label{fig:example_of_golden_roadmap}
\end{figure*}

We present a Golden Roadmap in Figure~\ref{fig:example_of_golden_roadmap}, where
nodes marked in pink font represent key nodes annotated by experts.

\begin{figure*}
  \begin{custombox}
    [Template for Extracting Core Research Problems and Skill Points from
    Dissertations.]
    \begin{lstlisting}
  You are a research assistant specialized in extracting key information from academic literature. Based on the provided dissertation, your task is to identify and extract core research problem and skill points. The requirements are:
  1. Extraction must be based strictly on the content of the provided paper.
  2. Extract at most five of the most innovative and valuable skill points.
  3. Extract the core research task of the entire paper.
  4. Answer in JSON format and follow the template provided below.
  5. Enclose the extracted results with ```json```.
  6. Answer in English.
  Template:
  {
    "core_research_question": ""
      # The core research question in a concise question format like "How to...?" (no more than 30 words).
      "skill_points": [
      {
      "problem_description": "", # Describe the specific problem this skill point addresses.
      "skill_point_name": "", # Concise name of the skill point.
      "skill_point_description": "" # Detailed explanation of the skill point content.
      },
      # Other skill points...
    ]
  }
  
    \end{lstlisting}
  \end{custombox}
  \caption{Template for extracting core research problems and skill points from
  dissertations.}
  \label{fig:template_for_extraction}
\end{figure*}

\begin{figure*}
    \begin{custombox}
        [Prompt of the $\mathcal{K}$ Agent for Generation of Internal Knowledge]
        \begin{lstlisting}
You are an experienced research expert, specialized in analyzing research problems and identifying key skills.
For a given research problem provided by the user, you need to analyze and output several key skills that are essential for solving the problem.

Please strictly follow the requirements below:
1. Analyze the research problem and output several key skills that are essential for solving the problem.
2. Output format requirements:
    - Use JSON list format for output, with each item being a skill point including name and description fields.
    - `name`: The name of the skill point (brief description).
    - `description`: The detailed explanation of the skill point (explains the role of the skill point in solving the problem).
    - Enclose the output in ```json```.
3. The skills should be answered in {split_language}.

Example output format:
```json
[
    {{"name": "xxx", "description": "xxx"}},
    {{"name": "xxx", "description": "xxx"}},
    ...
]
```
      \end{lstlisting}
    \end{custombox}
    \caption{ Prompt of the $\mathcal{K}$ agent for generation of internal knowledge.
    }
    \label{fig:prompt_of_K}
\end{figure*}

\begin{figure*}
    \begin{custombox}
        [Prompt of the $\mathcal{K}$ Agent for Knowledge Augmentation] \begin{lstlisting}
You are an experienced research expert, specialized in optimizing research problem roadmaps based on skill point repositories.
For a research problem, a current roadmap (in Markdown format), and a skill point repository provided by the user, you need to optimize the roadmap based on the research problem, the current roadmap, and the skill point repository.

Please strictly follow the requirements below:
1. Carefully analyze each skill point in the skill point repository (each skill point includes its name and description) according to the research problem and the current roadmap.
2. Determine which skill points are helpful for improving the roadmap, insert the names of helpful skill points as a step node into appropriate positions in the roadmap (names can be adapted to the problem as needed), and ignore unhelpful skill points.
3. Ensure the correct format of the roadmap during insertion, maintaining the logical order and continuity of node indices.
4. Roadmap format requirements, make sure to strictly follow:
    - Use Markdown format for output, with each line as a node (including level, index, and title, separated by spaces).
    - Use different heading levels (#, ##, ###, etc.) to indicate the node level and the hierarchical structure of the roadmap.
    - Use indices like 1.1.1 to indicate the node's position in the roadmap.
    - The title should use the format [xxx], where xxx is the node content.
5. Enclose the output in ```markdown```.
6. The roadmap content should be answered in {split_language}.

Example output format:
```markdown
# 1 [Main Step]
## 1.1 [Sub-step]
## 1.2 [Sub-step]
# 2 [Main Step]
## 2.1 [Sub-step]
### 2.1.1 [Sub-step]
### 2.1.2 [Sub-step]
...
```
      \end{lstlisting}
    \end{custombox}
    \caption{Prompt of the $\mathcal{K}$ agent for knowledge augmentation.}
    \label{fig:prompt_of_K_for_augmentation}
\end{figure*}

\begin{figure*}
    \begin{custombox}
        [Prompt of the $\mathcal{E}$ Agent] \begin{lstlisting}
You are an experienced research expert, specialized in evaluating the quality of research roadmaps.
For a research problem and a proposed roadmap (in Markdown format) used to solve the research problem, you need to evaluate the roadmap from multiple dimensions below and provide objective and fair scores and detailed analysis:
1. Logic Structure: Evaluate the logical coherence of the roadmap, including whether the dependencies between nodes are reasonable, whether the step order is logical, and whether there are conflicts or contradictions between different parts.
2. Granularity Degree: Evaluate the rationality of task decomposition granularity, including whether there are too macroscopic or too microscopic nodes, whether the sub-task division is balanced, etc.
3. Topic Relevance: Evaluate the relevance of the roadmap to the input research problem, including whether the roadmap fully covers the core content required to solve the problem, whether there are redundant nodes unrelated to the topic, and whether the key technical points are fully reflected, etc.
4. Completeness: Evaluate the richness and completeness of the roadmap content, including whether there are missing key nodes or steps, and whether all the content required to solve the problem is fully covered.

Based on the four dimensions above, provide an overall assessment using the following 100-point scale (percentage system, 0-100):
0-19: Very poor quality, cannot be used.
20-39: Poor quality, needs significant improvement and is not recommended as is.
40-59: Acceptable but still needs improvement before use.
60-79: Good quality and can be used with minor adjustments.
80-100: Excellent quality that are perfectly perfect in all four dimensions and can be directly used.

Output Format Requirements:
You must answer in English and enclose your score within <eval_score> tags and your detailed reasoning within <eval_reason> tags. For instance:

<eval_score>68</eval_score>
<eval_reason>Your detailed analysis covering all four dimensions...</eval_reason>

Please strictly follow the requirements above and provide the output in the specified format.
      \end{lstlisting}
    \end{custombox}
    \caption{Prompt of the $\mathcal{E}$ agent.}
    \label{fig:prompt_of_E}
\end{figure*}

\end{document}